\documentclass[graybox]{svmult}

\usepackage{mathptmx}       
\usepackage{helvet}         
\usepackage{courier}        
\usepackage{type1cm}
\usepackage{bm}
\usepackage{makeidx}         
\usepackage{graphicx}        
\usepackage{multicol}        
\usepackage[bottom]{footmisc}

\usepackage[utf8]{inputenc}
\usepackage[english]{babel}
\usepackage{fontenc}

\usepackage[round]{natbib} 
\usepackage{url}

\usepackage{amssymb,amsmath}

\renewcommand{\cite}[1]{\citep{#1}}
\providecommand{\textcite}[1]{\citeauthor{#1} \citeyearpar{#1}}

\setcitestyle{square}

\DeclareSymbolFont{matha}{OML}{txmi}{m}{it}
\DeclareMathSymbol{\varv}{\mathord}{matha}{118}

\newcommand{\og}[1]{''}
\newcommand{\fg}[1]{''}

\definecolor{orange}{RGB}{255,127,0}
\newcommand{\PW}[1]{{\color{black}#1}}

\usepackage[normalem]{ulem}
\usepackage{xcolor,cancel}

\newcommand{\States}{\mathbf{S}}
\newcommand{\State}{S}
\newcommand{\state}{s}
\newcommand{\Actions}{\mathbf{A}}
\newcommand{\Action}{A}
\newcommand{\action}{a}
\newcommand{\Transitions}{T}
\newcommand{\Rewards}{R}
\newcommand{\reward}{r}
\newcommand{\horizon}{H}
\newcommand{\history}{h}
\newcommand{\vf}{v}
\renewcommand{\varv}{\vf}

\newcommand{\argmax}{\operatorname*{argmax}}
\newcommand{\argmin}{\operatorname*{argmin}}

\makeindex             
                       
\begin{document}
                       
\title*{Reinforcement Learning}\label{volume-1-chapitre-RL}
\author{Olivier Buffet \and Olivier Pietquin \and Paul Weng}
\institute{Olivier Buffet \at INRIA, Universit\'{e} de Lorraine, CNRS, UMR 7503 - LORIA, Nancy, France\\ \email{olivier.buffet@loria.fr} \and
Olivier Pietquin \at Univ. Lille, CNRS, Centrale Lille, Inria, UMR 9189 - CRIStAL, Lille; France,  now with DeepMind, London, UK\\ \email{olivier.pietquin@univ-lille1.fr} \and
Paul Weng \at Shanghai Jiao Tong University, University of Michigan-Shanghai Jiao Tong University Joint Institute, Shanghai, China\\ \email{paul.weng@sjtu.edu.cn}}
\maketitle

\abstract{
Reinforcement learning (RL) is a general framework for adaptive control, which has proven to be efficient in many domains, e.g., board games, video games or autonomous vehicles.
In such problems, an agent faces a sequential decision-making problem
where, at every time step, it observes its state, performs an action, receives a reward and moves to a new state.
An RL agent learns by trial and error a good policy (or controller) based on observations and numeric reward feedback on the previously performed action.
In this chapter, we present the basic framework of RL and recall the two main families of approaches that have been developed to learn a good policy.
The first one, which is value-based, consists in  estimating the value of an optimal policy, value from which a policy can be recovered, while the other, called policy search, directly works in a policy space. 
Actor-critic methods can be seen as a policy search technique where the policy value that is learned guides the policy improvement.
Besides, we give an overview of some extensions of the standard RL framework, notably when risk-\PW{averse} behavior needs to be taken into account or when rewards are not available or not known.
}

\section{Introduction}

Reinforcement learning (RL) is a general framework for building autonomous agents (physical or virtual), which are systems that make decisions without human supervision in order to perform a given task.
Examples of such systems abound: expert backgammon player \cite{Tesauro95}, dialogue systems \cite{SinghKearnsLitmanWalker99}, 
acrobatic helicopter flight \cite{AbbeelCoatesNg10}, human-level video game player \cite{MnihKavukcuogluSilverRusuVenessBellemareGravesRiedmillerFidjelandOstrovskiPetersenBeattieSadikAntonoglouKingKumaranWierstraLeggHassabis15}, go player \cite{SilverHuangMaddisonGuezSifreDriesscheSchrittwieserAntonoglouPanneerschelvamLanctotDielemanGreweNhamKalchbrennerSutskeverLillicrapLeachKavukcuogluGraepelHassabis16} or autonomous driver \cite{BojarskiTestaDworakowskiFirnerFleppGoyalJackelMonfortMullerZhangZhangZhao16}. 
See also Chapter 11 of Volume 2 and Chapters 10 and 12 of Volume 3.

In all those examples, an agent faces a sequential decision-making problem, which can be represented as an interaction loop between an agent and an environment.
After observing its current situation, the agent selects an action to perform.
As a result, the environment changes its state and provides a numeric reward feedback about the chosen action.
In RL, the agent needs to learn how to choose good actions based on its observations and the reward feedback, without necessarily knowing the dynamics of the environment.

In this chapter, we focus on the basic setting of RL that assumes a single learning agent with full observability.
Some work has investigated the partial observability case (see \cite{Spaan12} for an overview of both the model-based and model-free approaches). 
The basic setting has also been extended to situations where several agents interact and learn simultaneously (see \cite{BusoniuBabuskaDeSchutter10} for a survey).
RL has also been tackled with Bayesian inference techniques, which we do not mention here for space reasons (see \cite{GhavamzadehMannorPineauTamar15} for a survey).

In Section~\ref{sec:basics}, we recall the Markov decision process model on which RL is formulated and the RL framework, along with some of their classic solution algorithms.
We present two families of approaches that can tackle large-sized problems for which function approximation is usually required.
The first, which is value-based, is presented in Section~\ref{sec:approx}.
It consists in estimating the value function of an optimal policy.
The second, called policy search, is presented in Section~\ref{sec:policysearch}.
It searches for an optimal policy directly in a policy space.
In Section~\ref{sec:extension}, we present some extensions of the standard RL setting, namely extensions to the case of unknown rewards and risk-sensitive RL approaches. 
Finally, we conclude in Section~\ref{sec:conclusion}.

\section{Background for RL}
\label{sec:basics}

Before presenting the RL framework, we recall the Markov decision process (MDP) model, on which RL is based. 
See also Chapter 17 of this volume and Chapter 10 of Volume 2.

{\em Markov decision process.} MDPs and their multiple variants (e.g., Partially Observable MDP or POMDP) \cite{Puterman94} 
have been proposed to represent and solve sequential decision-making problems under uncertainty. 
An MDP is defined as a tuple $\mathcal M = \langle \States, \Actions, \Transitions, \Rewards, \gamma, \horizon\rangle$ where $\States$ is a set of states, $\Actions$ is a set of actions, transition function $\Transitions(\state, \action, \state')$ specifies the probability of reaching state $\state'$ after performing action $\action$ in state $\state$, reward function $\Rewards(\state, \action) \in \mathbb R$ yields the immediate reward after performing action $\action$ in state $\state$, $\gamma \in [0, 1]$ is a discount factor and $\horizon \in \mathbb{N} \cup \{\infty\}$ is the horizon of the problem, which is the number of decisions to be made.
An immediate reward, which is a scalar number, measures the value of performing an action in a state. 
In some problems, it can be randomly generated.
In that case, $\Rewards(\state, \action)$ is simply the expectation of the random rewards.
In this MDP formulation, the environment is assumed to be stationary. 
Using such an MDP model, a system designer needs to define the tuple $\mathcal M$ such that an optimal policy performs the task s/he wants.

Solving an MDP (i.e., {\em planning}) amounts to finding 
a controller, called a {\em policy}, which specifies which action to take in every state of the environment in order to maximize the expected discounted sum of rewards (standard decision criterion).
A policy $\pi$ can be deterministic (i.e., $\pi(\state) \in \Action$) or randomized (i.e., $\pi(\cdot \,|\, \state)$ is a probability distribution over $\Actions$).
It can also be stationary or time-dependent, which is useful in finite-horizon or non-stationary problems.

A $t$-step history (also called trajectory, rollout or path) $\history = (\state_1, \action_1, \state_2, \ldots, \state_{t+1}) \in (\States\times\Actions)^t \times \States$ is a sequence of past states and actions. 
In the standard case, it is valued by its return defined as $\sum_t \gamma^{t-1}\Rewards(\state_t, \action_t)$.
As a policy induces a probability distribution over histories,
the {\em value function} $\vf^\pi : \States \to \mathbb{R}$ of a policy $\pi$ is defined by:
\begin{align*}
\vf^\pi_\horizon(\state) = \mathbb{E}_{\pi}\big[ \sum_{t=1}^\horizon \gamma^{t-1}\Rewards(\State_t, \Action_t) \,|\, \State_1 = \state \big],
\end{align*}
where $\mathbb{E}_{\pi}$ is the expectation with respect to the distribution induced by $\pi$ in the MDP, and $\State_t$ and $\Action_t$ are random variables respectively representing a state and an action at a time step $t$.
We will drop subscript $\horizon$ if there is no risk of confusion.
The value function can be computed recursively.
For deterministic policy $\pi$, we have:
\begin{align*}
\vf^\pi_0(\state) & = 0,\\
\vf^\pi_t(\state) & = \Rewards(\state, \pi(\state)) + \gamma \sum_{\state'\in\States} \Transitions(\state, \pi(\state), \state') \vf^\pi_{t-1}(\state').
\end{align*}
In a given state, policies can be compared via their value functions.
Interestingly, in standard MDPs, there always exists an optimal deterministic policy whose value function is maximum in every state.
Its value function is said to be optimal.

In the infinite horizon case, when $\gamma<1$, $\vf^\pi_t$ is guaranteed to converge to $\vf^\pi$, which is the solution of the {\em Bellman evaluation equations}:
\begin{align}
\vf^\pi(\state) & = \Rewards(\state, \pi(\state)) + \gamma \sum_{\state'\in\States} \Transitions(\state, \pi(\state), \state') \vf^\pi(\state'). \label{eq:bellmanevaluation}
\end{align}
Given $\vf^\pi$, a better policy can be obtained with the following improvement step:
\begin{align}
\pi'(\state) = \argmax_{\action \in \Actions} \Rewards(\state, \action) + \gamma \sum_{\state'\in\States} \Transitions(\state, \action, \state') \vf^\pi(\state'). \label{eq:improvement}
\end{align}
The policy iteration algorithm consists in alternating between a policy evaluation step (\ref{eq:bellmanevaluation}) and a policy improvement step (\ref{eq:improvement}), which converges to the optimal value function $\vf^* : \States \to \mathbb R$.

Alternatively, the optimal value function $\vf^*_\horizon : \States \to \mathbb{R}$ can also be iteratively computed for any horizon $\horizon$ by:
\begin{align}
\vf^*_0(\state) & = 0 \nonumber\\
\vf^*_t(\state) & = \max_{\action \in \Actions}\Rewards(\state, \action) + \gamma \sum_{\state'\in\States} \Transitions(\state, \action, \state') \vf^*_{t-1}(\state'). \label{eq:vi}
\end{align}
In the infinite horizon case, when $\gamma<1$, $\vf^*_t$ is guaranteed to converge to $\vf^*$, which is the solution of the {\em Bellman optimality equations}:
\begin{align}
\vf^*(\state) & = \max_{\action \in \Actions}\Rewards(\state, \action) + \gamma \sum_{\state'\in\States} \Transitions(\state, \action, \state') \vf^*(\state'). \label{eq:bellmanoptimality}
\end{align}
In that case, (\ref{eq:vi}) leads to the value iteration algorithm.

Two other related functions are useful when solving an RL problem:
the action-value function $Q^\pi_t(\state, \action)$ (resp. the optimal action-value function $Q^*_t(\state, \action)$) specifies the value of choosing an action $\action$ in a state $\state$ at time step $t$ and assuming policy $\pi$ (resp. an optimal policy) is applied thereafter, i.e.,
\begin{align*}
Q^x_t(\state, \action) = \Rewards(\state, \action) + \gamma \sum_{\state'\in\States} \Transitions(\state, \action, \state') \vf^x_{t-1}(\state') \quad \mbox{ where } x \in \{\pi, *\}. 
\end{align*}

{\em Reinforcement learning.} In the MDP framework, a complete model of the environment is assumed to be known (via the transition function) and the task to be performed is completely described (via the reward function). 
The RL setting has been proposed to tackle situations when those assumptions do not hold. 
An RL agent searches for (i.e., during the {\em learning phase}) a best policy while interacting with the unknown environment by trial and error.
In RL, the standard decision criterion used to compare policies is the same as in the MDP setting.
Although the reward function is supposed to be unknown, the system designer has to specify it.

In RL, value and action-value functions have to be estimated.
For $\vf^\pi$ of a given policy $\pi$, this can be done with the standard TD(0) evaluation algorithm, where the following update is performed after applying $\pi$ in state $\state$ yielding reward $\reward$ and moving to new state $\state'$:
\begin{align}
\vf^\pi_t(\state) = \vf^\pi_{t-1}(\state) - \alpha_t(\state) \left( \vf^\pi_{t-1}(\state) - \left(\reward + \vf^\pi_{t-1}(\state')\right)\right),\label{eq:td update rule}
\end{align}
where $\alpha_t(\state) \in [0, 1]$ is a learning rate.
For $Q^\pi$, the update is as follows, after the agent executed action $\action$ in state $\state$, received $\reward$, moved to new state $\state'$ and executed action $\action'$ (chosen by $\pi$):
\begin{align}
Q^\pi_t(\state, \action) = Q^\pi_{t-1}(\state, \action) - \alpha_t(\state, \action) \big( Q^\pi_{t-1}(\state, \action) - \big(\reward + \gamma Q^\pi_{t-1}(\state', \action')\big) \big),\label{eq:sarsa update rule}
\end{align}
where $\alpha_t(\state, \action) \in [0, 1]$ is a learning rate.
This update leads to the SARSA algorithm (named after the variables $\state, \action, \reward, \state', \action'$).
In the same way that the policy iteration algorithm alternates between an evaluation step and a policy improvement step, one can use the SARSA evaluation method and combine it with a policy improvement step. 
In practice, we do not wait for the SARSA evaluation update rule to converge to the actual value of the current policy to make a policy improvement step. 
We rather continuously behave according to the current estimate of the $Q$-function to generate a new transition. 
One common choice is to use the current estimate in a softmax (Boltzmann) function of temperature $\tau$ and behave according to a randomized policy: 
\[\pi_t(\action \,|\,\state)=\frac{e^{Q_{\theta_t}(\state, \action)/\tau}}{\sum_b e^{Q_{\theta_t}(\state, b)/\tau}}.\]

Notice that we chose to use the Bellman evaluation equations to estimate the targets. 
However we could also use the Bellman optimality equations in the case of the $Q$-function and replace $\reward + \gamma Q(\state', \action')$ by $\reward + \max_b Q(\state',b)$. 
Yet this only holds if we compute the value $Q^*$ of the optimal policy $\pi^*$. 
This gives rise to the $Q$-learning update rule, which directly computes the value of the optimal policy. 
It is called an {\em off-policy} algorithm (whereas SARSA is {\em on-policy}) because it computes the value function of another policy than the one that selects the actions and generates the transitions used for the update. 
The following update is performed after the agent executed action $\action$ (e.g., chosen according to the softmax rule) in state $\state$, received $\reward$ and moved to new state $\state'$:
\begin{align}
Q^*_t(\state, \action) = Q^*_{t-1}(\state, \action) - \alpha_t(\state, \action) \big( Q^*_{t-1}(\state, \action) - (\reward + \gamma \max_{\action'} Q^*_{t-1}(\state', \action')) \big) .\label{eq:qlearning update rule}
\end{align}

Updates (\ref{eq:td update rule}), (\ref{eq:sarsa update rule}) and (\ref{eq:qlearning update rule}) can be proved to converge if the learning rates satisfy standard stochastic approximation conditions (i.e., $\sum_t \alpha_t = \infty$ and $\sum_t \alpha_t^2 < \infty$).
Besides, for (\ref{eq:sarsa update rule}), 
temperature $\tau$ would also need to converge to $0$ while ensuring sufficient exploration in order for SARSA to converge to the optimal Q-function.
In practice, $\alpha_t(s, a)$ is often chosen constant, which would also account for the case where the environment is non-stationary.

Those two general framework (MDP and RL) have been successfully applied in many different domains. 
For instance, MDPs or their variants have been used in 
finance \cite{BauerleRieder11} or 
logistics \cite{ZhaoChenLeungLai10}.
RL has been applied to 
soccer \cite{BaiWuChen13} or 
power systems \cite{YuZhang13}, to cite a few. 
To tackle real-life large-sized problems, MDP and RL have to be completed with other techniques, such as  
compact representations \cite{BoutilierDeardenGoldszmidt00,GuestrinHauskrechtKveton04,Otterlo09} or 
function approximation \cite{de-FariasVan-Roy03,GeistPietquin11,MnihKavukcuogluSilverRusuVenessBellemareGravesRiedmillerFidjelandOstrovskiPetersenBeattieSadikAntonoglouKingKumaranWierstraLeggHassabis15}.

\section{Value-Based Methods with Function Approximation}\label{sec:approx}

In many cases, the state-action space is too large so as to be able to represent exactly the value functions $\vf^\pi$ or the action-value function $Q^\pi$ of a policy $\pi$. 
For this reason, function approximation for RL has been studied for a long time, starting with the seminal work of~\citet{BellmanPoly}. 
In this framework, the functions are parameterized by a vector of $d$ parameters $\bm\theta=[\theta_j]_{j=1}^{d}$, with $\bm\theta \in \Theta \subset \mathbb{R}^d$ (we will always 
consider column vectors) and the algorithms will aim at learning the parameters from data provided in the shape of transitions $\{\state_t, \action_t, \state'_t, \reward_t\}_{t=1}^N$ where $\state'_t$ is the successor state of $\state_t$ drawn from $\Transitions(\state_t, \action_t, \cdot)$. 
We will denote the parameterized versions of the functions as $\vf_{\bm\theta}$ and $Q_{\bm\theta}$. Popular approximation schemes are linear function approximation and neural networks. The former gave birth to a large literature in the theoretical domain as it allows studying convergence rates and bounds (although it remains non-trivial). 
The latter, although already used in the 90's~\cite{Tesauro95}, has known a recent growth in interest following the Deep Learning successes in supervised learning. 

The case of neural networks will be addressed in Section~\ref{sec:deep} but we will start with linear function approximation. In this particular case, a set of basis functions $\bm\phi(\cdot) = [\phi_j(\cdot)]_{j=1}^{d}$ has to be defined by the practitioner (or maybe learned through unsupervised learning) so that the value functions can be approximated by:
$$\vf_{\bm\theta}(\state) = \sum_j \theta_j \phi_j(\state) = \bm\theta^\intercal \bm\phi(\state) \quad \text{ or } \quad Q_{\bm\theta}(\state, \action) = \sum_j \theta_j \phi_j(\state, \action) = \bm\theta^\intercal \bm\phi(\state, \action).$$
The vector space defined by the span of $\bm\phi$ is denoted $\Phi$.

Notice that the exact case in which the different values of the value functions can be stored in a table (tabular case) is a particular case of linear function approximation. 
Indeed, if we consider that the state space is finite and small $\left( \state =\{\state_k\}_{k=1}^{|\States|} \in \States \right)$, then the value function can be represented in a table of $|\States|$ values 
$\{v_k \,|\, v_k = \vf(\state_k)\}_{k=1}^{|\States|}$ where $|\States|$ is the number of states. 
This is equivalent to defining a vector of $|\States|$ parameters $\bm{\vf}=[\vf_k]_{k=1}^{|\States|}$ and a vector of $|\States|$ basis functions $\bm\delta(\state)=[\delta_k(\state)]_{k=1}^{|\States|}$ where $\delta_k(\state)=1$ if $\state = \state_k$ and $0$ otherwise. 
The value function can thus be written $\vf(\state)=\sum_k \vf_k \delta_k(\state) = \bm\vf^\intercal \bm\delta(\state)$. 

\subsection{Stochastic Gradient Descent Methods}
\subsubsection{Bootstrapped Methods}\label{sec:bootstrap}

If one wanted to cast the Reinforcement Learning problem into a supervised learning problem (see Chapter 11 of this Volume and Chapter 12 of Volume 2), one could want to fit the parameters to the value function directly. For instance, to evaluate the value of a particular policy $\pi$, one would solve the following regression problem (for some $\ell_p$-norm and distribution $\mu$ over states):
$$\bm\theta^* = \argmin_{\bm\theta} \|\vf^\pi_{\bm\theta} - \varv^\pi \|_{p, \mu} = \argmin_{\bm\theta} \|\vf^\pi_{\bm\theta} - \varv^\pi \|_{p, \mu}^p$$
where $\|\cdot\|_{p, \mu}$ \PW{denotes the weighted $\ell_p$-norm defined by $ \big(\mathbb{E}_\mu |\cdot|^p\big)^{1/p}$}, $\mathbb{E}_\mu$ is the expectation with respect to $\mu$\PW{}.
Yet, as said before, we usually cannot compute these values everywhere and we usually only have access to some transition samples $\{\state_t, \action_t, \state'_t, \reward_t\}_{t=1}^N$ generated according to distribution $\mu$. 
So we could imagine casting the RL problem into the following minimization problem:
$$\bm\theta^* = \argmin_{\bm\theta} \frac{1}{N} \sum_{t=1}^N \PW{|\vf_{\bm\theta}^\pi(\state_t) - \varv^\pi(\state_t) |^p}.$$
This cost function can be minimized by stochastic gradient descent (we will consider an $\ell_ 2$-norm):
\begin{align*}
\bm\theta_t &= \bm\theta_{t-1} - \frac{\alpha}{2} \nabla_{\bm\theta_{t-1}} \PW{\big(\vf_{\bm\theta_{t-1}}^\pi(\state_t) - \varv^\pi(\state_t)\big)^2} \\
& = \bm\theta_{t-1} - \alpha \nabla_{\bm\theta_{t-1}} \vf_{\bm\theta_{t-1}}^\pi(\state_t) \left(\vf_{\bm\theta_{t-1}}^\pi(\state_t) - \varv^\pi(\state_t) \right).
\end{align*}

Of course, it is not possible to apply this update rule as it is since we do not know the actual value $\varv^\pi(\state_t)$ of the states we observe in the transitions. 
But, from the Bellman evaluation equations (\ref{eq:bellmanevaluation}), we can obtain an estimate by replacing $\varv^\pi(\state_t)$ by $\reward_t + \gamma \vf^\pi_{\bm\theta_{t-1}}(\state_{t+1})$. 
Notice that this replacement uses bootstrapping as we use the current estimate of the target to compute the gradient. 
We finally obtain the following update rule for evaluating the current policy $\pi$:
$$\bm\theta_t = \bm\theta_{t-1} - \alpha \nabla_{\bm\theta_{t-1}} \vf_{\bm\theta_{t-1}}^\pi(\state_t) \left(\vf_{\bm\theta_{t-1}}^\pi(\state_t) - \left(\reward_t + \gamma \vf^\pi_{\bm\theta_{t-1}}(\state'_{t})\right)\right).$$
In the case of linear function approximation, i.e., $\vf^\pi_{\bm\theta}(\state) = \bm\theta^\intercal \bm\phi(\state)$, we obtain:
$$\bm\theta_t = \bm\theta_{t-1} - \alpha \bm\phi(\state_t) \left(\bm\theta_{t-1}^\intercal \bm\phi(\state_t) - \left(\reward_t + \gamma \bm\theta_{t-1}^\intercal \bm\phi(\state'_{t})\right)\right).$$
Everything can be written again in the case of the action-value function, which leads to the SARSA update rule with linear function approximation $Q^\pi_{\bm\theta}(\state, \action) = \bm\theta^\intercal \bm\phi(\state, \action)$: 
$$\bm\theta_t = \bm\theta_{t-1} - \alpha \bm\phi(\state_t, \action_t) \left( \bm\theta_{t-1}^\intercal \bm\phi(\state_t, \action_t) - \left(\reward_t + \gamma \bm\theta_{t-1}^\intercal  \bm\phi(\state'_{t}, \action'_{t})\right)\right).$$
Changing the target as in the Q-learning update, we obtain for $Q^*_{\bm\theta}(\state, \action) = \bm\theta^\intercal \bm\phi(\state, \action)$:
$$\bm\theta_t = \bm\theta_{t-1} - \alpha \bm\phi(\state_t, \action_t) \left( \bm\theta_{t-1}^\intercal \bm\phi(\state_t, \action_t) - \left(\reward_t + \gamma \max_b \bm\theta_{t-1}^\intercal \bm\phi(\state'_{t},b)\right)\right).$$

\subsubsection{Residual Methods}\label{sec:residual}

Instead of using the Bellman equations to provide an estimate of the target after deriving the update rule, one could use it directly to define the loss function to be optimized. We would then obtain the following minimization problem:
$$\bm\theta^* = \argmin_{\bm\theta} \frac{1}{N} \sum_{t=1}^N \PW{\big( \vf_{\bm\theta}^\pi(\state_t) - \left(\reward_t + \gamma \vf^\pi_{\bm\theta} (\state'_{t})\right) \big)^2}.$$
This can also be seen as the minimization of the Bellman residual. 
Indeed the Bellman evaluation equations 
($\vf^\pi(\state) = \mathbb{E}_\pi[\Rewards(\state, \Action)+\gamma \vf^\pi(\State')]$) can be rewritten as 
$\vf^\pi(\state) - \mathbb{E}_\pi[\Rewards(\state, \Action)+\gamma \vf^\pi(\State')] = 0$. 
So by minimizing the quantity $\vf^\pi(\state) - \mathbb{E}_\pi[\Rewards(\state, \Action)+\gamma \vf^\pi(\State')]$, called the Bellman residual, we reach the objective of evaluating $\vf^\pi(\state)$. 
Here, we take the observed quantity $\reward+\gamma \vf^\pi(\state')$ as an unbiased estimate of its expectation. 
The Bellman residual can also be minimized by stochastic gradient descent as proposed by~\citet{baird1995residual} and the update rule becomes:
$$\bm\theta_t = \bm\theta_{t-1} - \alpha \nabla_{\bm\theta_{t-1}} \left( \vf_{\bm\theta_{t-1}}^\pi(\state_t) - \left( \reward_t + \gamma \vf_{\bm\theta_{t-1}}^\pi (\state'_{t})\right) \right) \left(\vf_{\bm\theta_{t-1}}^\pi(\state_t) - \left(\reward_t + \gamma \vf^\pi_{\bm\theta_{t-1}}(\state'_{t})\right)\right).$$
In the case of a linear approximation, we obtain:
$$\bm\theta_t = \bm\theta_{t-1} - \alpha \left( \bm\phi(s_t) - \gamma \bm\phi(\state'_{t}) \right) \left( \bm\theta_{t-1}^\intercal \bm\phi(\state_t) - \left(\reward_t + \gamma \bm\theta_{t-1}^\intercal \bm\phi(\state'_{t})\right)\right).$$
This approach, called R-SGD (for residual stochastic gradient descent), has a major flaw as it computes a biased estimate of the value-function. 
Indeed, $\vf_{\bm\theta}^\pi(\state_t)$ and $\vf_{\bm\theta}^\pi(\state'_t)$ are correlated as $\state'_t$ is the result of having taken action $\action_t$ chosen by $\pi(\state_t)$ \cite{Werbos90}. 
To address this problem,~\citet{baird1995residual} suggest to draw two different next states $\state'_{t}$ and $\state''_{t}$ starting from the same state $\state_t$ and to update as follows: 
$$\bm\theta_t = \bm\theta_{t-1} - \alpha \nabla_{\bm\theta_{t-1}} \left( \vf_{\bm\theta_{t-1}}^\pi(\state_t) - \left( \reward_t + \gamma \vf_{\bm\theta_{t-1}}^\pi (\state'_{t})\right) \right) \left(\vf_{\bm\theta_{t-1}}^\pi(\state_t) - \left(\reward_t + \gamma \vf^\pi_{\bm\theta_{t-1}}(\state''_{t})\right)\right).$$
Of course, this requires that a generative model or a simulator is available and that transitions can be generated on demand. 

The same discussions as in previous section can apply to learning an action-value function. 
For instance, one could want to solve the following optimization problem to learn the optimal action-value function: 
\begin{align}
\bm\theta^* = \argmin_{\bm\theta} \frac{1}{N} \sum_{t=1}^N \PW{\left( Q_{\bm\theta}^*(s_t, a_t) - \big(\reward_t + \gamma \max_b Q^*_{\bm\theta} (\state'_{t}, b)\big) \right)^2}. \label{eq:qres}
\end{align}

Yet this optimal residual cannot directly be minimized in the case of the $Q$-function as the $\max$ operator is not differentiable. Notice that a sub-gradient method can still be used.  

\subsection{Least-Squares Methods}

Gradient descent was used to minimize the empirical norm of either the bootstrapping error or the Bellman residual in the previous section. 
As the empirical norm is generally using the $\ell_2$-norm and that linear function approximation is often assumed, another approach could be to find the least squares solution to these problems. 
Indeed, least squares is a powerful approach as it is a second-order type of method and offers a closed-form solution to the optimization problem. 
Although there is no method that explicitly applies least squares to the two aforementioned empirical errors, one can see the fixed-point Kalman Filter (FPKF) algorithm~\cite{choi2006generalized} as a recursive least squares method applied to the bootstrapping error minimization. 
Also, the Gaussian Process Temporal Difference (GPTD)~\cite{GPTD} or the Kalman Temporal Difference (KTD)~\cite{KTD} algorithms can be seen as recursive least squares methods applied to Bellman residual minimization. 
We invite the reader to refer to~\citet{geist2013algorithmic} for further discussion on this. 

Yet, the most popular method inspired by least squares optimization does apply to a different cost function. The Least-Squares Temporal Difference (LSTD) algorithm~\citep{bradtke1996linear} aims at minimizing: 
$$\bm\theta^* = \argmin_{\bm\theta} \frac{1}{N} \sum_{i=1}^N \left(\vf^\pi_{\bm\theta}(\state_i) - \vf^\pi_{\bm\omega^*}(\state_i) \right)^2,$$
where $\bm\omega^* = \argmin_{\bm\omega} \frac{1}{N} \sum_{i=1}^N \left(\vf^\pi_{\bm\omega}(\state_i) - \left(\reward_i + \gamma \vf^\pi_{\bm\theta}(\state'_i) \right)\right)^2$ can be understood as a projection
on the space $\Phi$ spanned by the family of functions $\bm\phi_j$'s used to approximate $\vf^\pi$. 
It can be seen as the composition of the Bellman operator and of a projection operator. 
This cost function is the so-called {\em projected Bellman residual}. 
When using linear function approximation, 
this optimization problem admits a closed-form solution: 
$$\bm\theta^* = \left[\sum_{i=1}^N \bm\phi(\state_i) \left[\bm\phi(\state_i) - \gamma \bm\phi(\state'_i)\right]^\intercal  \right]^{-1} \sum_{i=1}^N \bm\phi(\state_i) \reward_i.$$ 
Note that the projected Bellman residual can also be optimized with a stochastic gradient approach \cite{SuttonMaeiPrecupBhatnagarSilverSzepesvariWiewiora09}.

Extensions to non-linear function approximation exist and rely on the kernel trick~\citep{xu2007kernel} or on statistical linearization~\cite{geist2010statistically}. 
LSTD can be used to learn an approximate $Q$-function as well and can be combined with policy improvement steps into an iterative algorithm, similar to policy iteration, to learn an optimal policy from a dataset of sampled transitions. 
This gives rise to the so-called Least Squares Policy Iteration (LSPI) algorithm \citep{lagoudakis2003least}, which is one of the most popular batch-RL algorithm. 

\subsection{Iterative Projected Fixed-Point Methods}

As we have seen earlier, dynamic programming offers a set of algorithms to compute value functions of a policy in the case the dynamics of the MDP is known. 
One of these algorithms, Value Iteration, relies on the fact that the Bellman equations define contraction operators when $\gamma<1$. 
For instance, if we define the Bellman evaluation operator $B^\pi$ such that $B^\pi Q(\state, \action) = \Rewards(\state, \action) + \gamma \mathbb{E}_{\pi}\big[Q(\State', \Action') \,|\, \State = \state, \Action=\action\big]$, one can show that iteratively applying $B^\pi$ to a random initialization of $Q$ converges to $Q^\pi$, because $B^\pi$ defines a contraction for which the only fixed point is $Q^\pi$~\citep{Puterman94}. 
The Bellman optimality operator $B^*$, defined as $B^*Q(\state, \action) = \Rewards(\state, \action) + \gamma \mathbb{E}\big[\max_b Q(\State', b) \,|\, \State = \state, \Action = \action\big]$, is also a contraction. 
The same holds for the sampled versions of the Bellman operators. 
For instance, let us define the sampled evaluation operator $\hat{B}^*$ such that $\hat{B}^* Q(\state, \action) = \reward + \gamma \max_b Q(\state', b)$, where the expectation has been removed (the sampled operator applies to a single transition). 
Unfortunately, there is no guarantee that this remains a contraction when the value functions are approximated. 
Indeed when applying a Bellman operator to an approximate $Q_{\bm\theta}$, the result might not lie in the space spanned by $\bm\theta$. 
One has thus to project back on the space $\Phi$ spanned by $\bm\phi$ using a projection operator $\Pi_\Phi$, i.e., 
$\Pi_\Phi f = \argmin_{\bm\theta} \| \bm\theta^\intercal \bm\phi - f \PW{\|_2} $. 
If the composition of $\Pi_\Phi$ and $\hat{B}^\pi$ (or $\hat{B}^*$) is still a contraction, then recursively applying this composition to any initialization of $\bm\theta$ still converges to a good approximate $Q^\pi_{\bm\theta}$ (or $Q^*_{\bm\theta}$). 
Unfortunately, the exact projection is often impossible to get as it is a regression problem. 
For instance, one would need to use least squares methods or stochastic gradient descent to learn the best projection from samples. 
Therefore the projection operator itself is approximated and will result in some $\hat{\Pi}_\Phi$ operator. 
So the iterative projected fixed-point process is defined as: 
$$ Q_{\bm\theta_t} = \hat{\Pi}_\Phi \hat{B}^\pi Q_{\bm\theta_{t-1}} \quad \text{ or } \quad Q_{\bm\theta_t} = \hat{\Pi}_\Phi \hat{B}^* Q_{\bm\theta_{t-1}}.$$

In practice, the algorithm consists in collecting transitions (e.g., $\{\state_i, \action_i, \reward_i, \state'_i\}_{i=1}^N $), initialize $\bm\theta_0$ to some random value, compute a regression database by applying the chosen sampled Bellman operator (e.g., $\{\hat{B}^*Q_{\bm\theta_0}(\state_i, \action_i) = \reward_i + \gamma \max_b Q_{\bm\theta_0}(\state_i, b)\}_{i=1}^N$), apply a regression algorithm to find the next value of parameters (e.g., $Q_{\bm\theta_1} = \hat{\Pi}_\Phi \hat{B}^* Q_{\bm\theta_{0}} = \argmin_{\bm\theta}
\frac{1}{N} \sum_{i=1}^N  \PW{\big(Q_{\bm\theta}(\state_i, \action_i) - \hat{B}^*Q_{\bm\theta_0}(\state_i, \action_i) \big)^2}$) and iterate. 

This method finds its roots in early papers on dynamic programming~\citep{samuel1959some,bellman1963polynomial} and convergence properties have been analyzed by
~\citet{gordon1995stable}. The most popular implementations use regression trees~\citep{ernst2005tree} or neural networks~\citep{riedmiller2005neural} as regression algorithms and have been applied to many concrete problems such as robotics~\citep{antos2008fitted}. 

\subsection{Value-Based Deep Reinforcement Learning}
\label{sec:deep}

Although the use of Artificial Neural Networks (ANN, see Chapter 12 of Volume 2) in RL is not new~\citep{Tesauro95}, there has been only a few successful attempts to combine RL and ANN in the past. 
Most notably, before the recent advances in Deep Learning (DL)~\citep{lecun2015deep}, one can identify the work by~\citet{riedmiller2005neural} as the biggest success of ANN as a function approximation framework for RL. 
There are many reasons for that, which are inherently due to the way ANN learns and assumptions that have to be made for both gradient descent and most value-based RL algorithms to converge. 
Especially, Deep ANNs (DNN) require a tremendous amount of data as they contain a lot of parameters to learn (typically hundreds of thousands to millions). 
To alleviate this issue, \PW{~\citet{Tesauro95}} trained his network to play backgammon through a self-play procedure. 
The model learned at iteration $t$ plays again itself to generate data for training the model at iteration $t+1$. 
It could thus reach super-human performance at the game of backgammon using RL. This very simple and powerful idea was reused in~\citep{SilverHuangMaddisonGuezSifreDriesscheSchrittwieserAntonoglouPanneerschelvamLanctotDielemanGreweNhamKalchbrennerSutskeverLillicrapLeachKavukcuogluGraepelHassabis16} to build the first artificial Go player that consistently defeated a human Go master. 
Yet, this method relies on the assumption that games can easily be generated on demand (backgammon and Go rules are simple enough even though the game is very complex). 
In more complex settings, the agent faces an environment for which it does not have access to the dynamics, maybe it cannot start in random states and has to follow trajectories, and it can only get transitions through actual interactions. 
This causes two major issues for learning with DNNs (in addition to intensive usage of data). 
First, gradient descent for training DNNs assume the data to be independent and identically distributed (i.i.d. assumption). 
Second, the distribution of the data should remain constant over time. 
Both these assumptions are normally violated by RL since transitions used to train the algorithms are part of trajectories (so next states are functions of previous states and actions, violating the i.i.d. assumption) and because trajectories are generated by a policy extracted from the current estimate of the value function (learning the value function influences the distribution of the data generated in the future). 
In addition, we also have seen in Section~\ref{sec:residual} that Bellman residual minimization suffers from the correlation between estimates of value functions of successive states. 
All these problems make RL unstable~\citep{gordon1995stable}.

To alleviate these  issues,~\citet{MnihKavukcuogluSilverRusuVenessBellemareGravesRiedmillerFidjelandOstrovskiPetersenBeattieSadikAntonoglouKingKumaranWierstraLeggHassabis15} used two tricks that allowed to reach super-human performances at playing Atari 2600 games from pixels. 
First, they made use of a biologically inspired mechanism, called experience replay~\citep{lin1992self}, that consists in storing transitions in a Replay Buffer $D$ before using them for learning. 
Instead of sequentially using these transitions, they are shuffled in the buffer and randomly sampled for training the network (which helps breaking correlation between successive samples). \PW{The buffer is filled on a first-in-first-out basis so that the distribution of the transitions is nearly stationary (transitions generated by old policies are discarded first).
Second, the algorithm is based on asynchronous updates of the network used for generating the trajectories and a slow learning network. 
The slow learning network, called the target network, will be updated less often than the network that actually learns from the transitions stored in the replay buffer (the $Q$-network). 
This way, }the update rule of the $Q$-network is built such that correlation between estimates of $Q(s,a)$ and $Q(s',a')$ is reduced. 
Indeed, the resulting algorithm (Deep Q-Network or DQN) is inspired by the gradient-descent update on the optimal Bellman residual (\ref{eq:qres}). 
But instead of using the double-sampling trick mentioned in Section~\ref{sec:residual}, two different estimates of the $Q$-function are used. 
One according to the target network parameters ($\bm\theta^-$) and the other according to $Q$-network parameters ($\bm\theta$). 
The parameters of the $Q$-network are thus computed as: 
$$\bm\theta^* = \argmin_{\bm\theta} \sum_{(\state_t, \action_t, \state'_t, \reward_t) \in D} \left[\left(\reward_t + \gamma \max_b Q_{\bm\theta^-}(\state_t', b)\right) - Q_{\bm\theta}(\state_t, \action_t) \right]^2,$$
With this approach, the problem of non-differentiability of the $\max$ operator is also solved as the gradient is computed w.r.t. $\bm\theta$ and not $\bm\theta^-$.
Once in a while, the target network parameters are updated with the $Q$-network parameters ($\bm\theta^- \leftarrow \bm\theta^* $) and new trajectories are generated according to the policy extracted from $Q_{\bm\theta^-}$ to fill again the replay buffer and train again the $Q$-network. 
The target network policy is actually a softmax policy based on $Q_{\bm\theta^-}$ (see Section~\ref{sec:bootstrap}). 
Many improvements have been brought to that method since its publication, such as a prioritized replay mechanism~\citep{schaul2016prioritized} that allows to sample more often from the replay buffer transitions for which the Bellman residual is larger, or the Double-DQN trick~\citep{van2016deep} used to provide more stable estimates of the $\max$ operator. 

\section{Policy-Search Approaches} \label{sec:policysearch}

Value-based approaches to RL rely on approximating the optimal value
function $V^*$ (typically using Bellman's optimality principle), and
then acting greedily with respect to this function.
Policy Search algorithms directly optimize control policies, which
\PW{typically} depend on a parameter vector $\bm\theta\in \Theta$ (and are thus
noted $\pi_{\bm\theta}$), and whose general shape is user-defined.%
\footnote{This section is mainly inspired by \cite{DeiNeuPet-ftr11},
  although that survey focuses on a robotic framework.}
Possible representations include linear policies, (deep) neural
networks, radial basis function networks, and dynamic movement
primitives (in robotics).
Using such approaches avoids issues with discontinuous value functions,
and makes it possible, in some cases, to deal with high-dimensional
(possibly continuous) state and action spaces.
They also allow providing expert knowledge through the shaping of the
controller, or through example trajectories---to initialize the
parameters.

In the following, we mainly distinguish between {\em model-free} and
{\em model-based} algorithms---\textit{i.e}., depending on whether a model is
being learned or not.

\subsection{Model-Free Policy Search}

In model-free policy search, sampled trajectories are used directly to
update the policy parameters.
The discussion will follow the three main steps followed by the
algorithms: %
(i) how they {\em explore} the space of policies, %
(ii) how they {\em evaluate} policies, and %
(iii) how policies are {\em updated}.

\subsubsection{Policy Exploration}

Exploring the space of policies implies either sampling the parameter
vector the policy depends on, or perturbing the action choice of the
policy.
Often, the sampling of parameters takes place at the beginning of each
episode (in episodic scenarios), and action perturbations are
different at each time step, but other options are possible.
Stochastic policies can be seen as naturally performing a step-based
exploration in action space.
Otherwise, the exploration strategy can be modeled as an {\em
  upper-level policy} $\pi_{\omega}(\theta)$---sampling $\theta$
according to a probability distribution governed by parameter vector
$\omega$---, while the actual policy $\pi_{\theta}(\action|\state)$ is
refered to as a {\em \PW{lower}-level policy}.
In this setting, the policy search aims at finding the parameter
vector $\omega$ that maximizes the expected return given this vector.
If $\pi_{\omega}(\theta)$ is a Gaussian distribution (common in
robotics), then its covariance matrix can be diagonal---typically in
step-based exploration---or not---which leads to more stability, but
requires more samples---, meaning that the various parameters in
$\theta$ can be treated in a correlated manner or not.

\subsubsection{Policy Evaluation}

Policy evaluation can also be step-based or episode-based.
Step-based approaches evaluate each state-action pair.
They have low variance and allow crediting several parameter vectors.
They can rely on $Q$-value estimates, which can be biased and prone to
approximation errors, or Monte-Carlo estimates, which can suffer from
high variance.
Episode-based approaches evaluate parameters using complete
trajectories.
They allow more performance criteria than step-based
approaches---\textit{e.g.}, minimizing the final distance to the target.
They also allow for more sophisticated exploration strategies, but
suffer all the more from noisy estimates and high variance that the
dynamics are more stochastic.

\subsubsection{Policy Update}

Finally, the policy can be updated in rather different manners.
We will discuss approaches relying on gradient ascents, inference-based
optimization, information-theoretic ideas, stochastic optimization and
path-integral optimal control.

{\bf Policy Gradient} (PG) algorithms first require estimating the
gradient.
Some (episode-based) PG algorithms perform this estimate using a
finite difference (FD) method by perturbing the parameter vector.
Other algorithms instead exploit the {\em
  Likelihood ratio} trick, which allows estimating the gradient from a
single trajectory, but requires a stochastic policy.
These can be step-based as REINFORCE \cite{Williams-ml92} or G(PO)MDP
\cite{BaxBar-jmlr01,BaxBarWea-jmlr01}, or episode-based as PEPG
\cite{SehEtAl-nn10}.

Policy gradients also include natural gradient algorithms (NPG), i.e.,
algorithms that try to limit the distance between distributions
$P_{\theta}(\history)$ and $P_{\theta+\delta\theta}(\history)$ using the KL
divergence (estimated through the Fisher information matrix (FIM)).
In step-based NPGs \cite{BagSch-ijcai03,PetSch-nn08}, using
appropriate (``{\em compatible}'') function approximation removes the
need to estimate the FIM, but requires estimating the value function,
which can be difficult.
On the contrary, episodic Natural Actor-Critic (eNAC)
\cite{PetSch-nc08} uses complete episodes, and thus only estimates
$\vf(\state_1)$.
NAC \cite{PetSch-nn08} addresses infinite horizon problems, the lack
of episodes leading to the use of Temporal Difference methods to
estimate values.

Policy gradient usually applies to randomized policies.
Recent work \cite{SilverLeverHeessDegrisWierstraRiedmiller14,LillicrapHuntPritzelHeessErezTassaSilverWierstra16} has adapted it to deterministic policies with a continuous action space, which can potentially facilitate the gradient estimation.
Building on DQN, actor-critic methods have been extended to asynchronous updates with parallel actors and neural networks as approximators \cite{MnihBadiaMirzaGravesLillicrapHarleySilverKavukcuoglu16}.

{\bf Inference-based algorithms} avoid the need to set learning rates.
They consider that (i) the return $R$ is an observed binary variable
($1$ meaning success),\footnote{Transformations can bring us in this
  setting.} (ii) the trajectory $\history$ is a latent variable, and (iii)
one looks for the parameter vector that maximizes the probability of
getting a return of $1$.
Then, an Expectation-Maximization algorithm can address this Bayesian
inference problem.
Variational inference can be used in the E-step of EM
\cite{Neumann-icml11}, but most approaches rely on Monte-Carlo
estimates instead, despite the fact that they perform maximum
likelihood estimates over several modes of the reward function (and
thus do not distinguish them).
These can be episode-based algorithms as RWR \cite{PetSch-esann07}
(uses a linear upper-level policy) or CrKR \cite{KobOztPet-rss10} (a
kernelized version of RWR, i.e., which does not need to specify
feature vectors, but cannot model correlations).
These can also be step-based algorithms as PoWER \cite{KobPet-ml10},
which allows a more structured exploration strategy, and gives more
influence to data points with less variance.

{\bf Information-theoretic} approaches (see Chapter 2 of Volume 3) try to limit changes in
trajectory distributions between two consecutive time steps, which
could correspond to degradations rather than improvements in the
policy.
Natural PGs have the same objective, but need a user-defined learning
rate. %
Instead, REPS \cite{PetMueAlt-aaai10} combines advantages from NPG
(smooth learning) and EM-based algorithms (no learning-rate). %
Episode-based REPS \cite{DanNeuPet-aistats12} learns a higher-level
policy while bounding parameter changes by solving a constrained
optimization problem. %
Variants are able to adapt to multiple contexts or learn multiple
solutions. %
Step-based REPS \cite{PetMueAlt-aaai10} solves an infinite horizon
problem (rather than an episodic one), optimizing the average reward
per time step. %
It requires enforcing the stationarity of state features, and thus
solving another constrained optimization problem. 
A related recent method, TRPO \cite{SchulmanLevineAbbeelJordanMoritz15}, which notably constrains the changes of $\pi(\cdot \,|\, \state)$ instead of those of state-action distributions, proves to work well in practice.

{\bf Stochastic Optimization} relies on black-box optimizers, and thus
can easily be used for episode-based formulations, i.e., working with
an upper-level policy $\pi_{\omega}(\theta)$. %
Typical examples are CEM \cite{BoeKroManRub-aor05,SziLor-nc05},
CMA-ES \cite{HanMulKou-ec03,HeiIge-joa09}, and
NES \cite{WierstraSchaulGlasmachersSunPetersSchmidhuber14}, three evolutionary algorithms
that maintain a parametric probability distribution (often Gaussian) $\pi_{\omega}(\theta)$ over the
parameter vector. %
They sample a population of candidates, evaluate them, and use the
best ones (weighted) to update the distribution. %
Many rollouts may be required for evaluation, as examplified with
the game of Tetris \cite{SziLor-nc05}.

{\bf Path Integral} (PI) approaches were introduced for optimal
control, i.e., to handle non-linear continuous-time systems.
They handle squared control costs and arbitrary state-dependent
rewards. %
{\em Policy Improvement with PIs} (PI$^2$) applies PI theory to
optimize Dynamic Movement Primitives (DMPs), i.e., representations of
movements with parameterized differential equations, using Monte-Carlo
rollouts instead of dynamic programming.

\subsection{Model-Based Policy Search}

Typical model-based policy-search approaches repeatedly %
(i) sample real-world trajectories using a fixed policy; %
(ii) learn a forward model of the dynamics based on these samples (and
previous ones); %
(iii) optimize this policy using the learned model (generally as a
simulator).
As can be noted, this process does not explicitly handle the
exploration-exploitation trade-off as policies are not chosen so as to
improve the model where this could be appropriate.
We now discuss three important dimensions of these approaches: how to
learn the model, how to make reliable long-term predictions, and how
to perform the policy updates.

Model learning often uses probabilistic models.
They first allow accounting for uncertainty due to sparse data
(at least in some areas) or an inappropriate model class.
In robotics, where action and state spaces are continuous,
non-parametric probabilistic methods can be used such as Linearly
Weighted Bayesian Regression (LWBR) of Gaussian Processes (GPs), which
may suffer from increasing time and memory requirements.
But probabilistic models can also be employed to represent stochastic
dynamics.
An example is that of propositional problems, which are often modeled
as Factored MDPs \cite{BouDeaGol-ijcai95}, where the dynamics and
rewards are DBNs whose structure is {\em a priori} unknown.
A variety of approaches have been proposed, which rely on different
representations (such as rule sets, decision trees, Stochastic STRIPS,
or PPDDL)
\cite{DegSigWui-icml06,PasZetKae-jair07,WalSziDiuLit-uai09,LesZan-ewrl11}.
See Chapter 10 of Volume 2.

Long-term predictions are usually required to optimize the policy
given the current forward model.
While the real world is its own best (unbiased) model, using a learned
model has the benefit of allowing to control these predictions.
A first approach, similar to paired statistical tests, is to always
use the same random initial states and the same sequences of random
numbers when evaluating different policies.
It has been introduced for policy-search in the PEGASUS framework
\cite{NgJor-uai00} and drastically reduces the sampling variance.
Another approach is, when feasible, to compute a probability
distribution over trajectories using deterministic approximations such
as linearization \cite{AndMoo-of05}, sigma-point methods (e.g., \citep{JulUhl-ieee04}) or moment-matching.

Policy updates can rely %
on gradient-free optimization (e.g., Nelder-Mead method or
hill-climbing) \citep{BagSch-icra01}, %
on sampling-based gradients (e.g., finite difference methods), as in model-free approaches,
although they require many samples, %
or on analytical gradients \citep{DeiRas-icml11}, which require the model as well as the
policy to be differentiable, scale favorably with the number of
parameters, but are computationally involved.

\section{Extensions: Unknown Rewards and Risk-sensitive Criteria} \label{sec:extension}

In the previous sections, we recalled different techniques for solving RL problems, with the assumption that policies are compared with the expected cumulated rewards as a decision criterion.
However, rewards may not be scalar, known or numeric, and the standard criterion based on expectation may not always be suitable.
For instance, multiobjective RL has been proposed to tackle situations where an action is evaluated over several dimensions (e.g., duration, length, power consumption for a navigation problem).
The interested reader may refer to \cite{RoijersVamplewWhitesonDazeley13} for a survey and refer to Chapter 16 of this volume for an introduction to multicriteria decision-making.
For space reasons, we focus below only on three extensions: reward learning (Section~\ref{sec:rewardlearning}), preference-based RL (Section~\ref{sec:preferencebased}) and risk sensitive RL (Section~\ref{sec:risksensitive}).

\subsection{Reward Learning} \label{sec:rewardlearning}

From the system designer's point of view, defining the reward function can be viewed as programming the desired behavior in an autonomous agent. 
A good choice of reward values may accelerate learning \cite{MatignonLaurentLe-Fort-Piat06} while an incorrect choice may lead to unexpected and unwanted behaviors \cite{RandlovAlstrom98}.
Thus, designing this function is a hard task (e.g., robotics \cite{ArgallChernovaVelosoBrowning09}, natural language parsers \cite{NeuSzepesvari09} or dialogue systems \cite{El-AsriLarochePietquin12}).

When the reward signal is not known, a natural approach is to learn from demonstration.
Indeed, in some domains (e.g., autonomous driving), it is much simpler for an expert to demonstrate how to perform a task rather than specify a reward function.
Such an approach has been called apprenticeship learning \cite{AbbeelNg04}, learning from demonstration \cite{ArgallChernovaVelosoBrowning09}, behavior cloning or imitation learning \cite{HusseinGaberElyanJayne17}. 
Two families of techniques have been developed to solve such problems.
The first group tries to directly learn a good policy from (near) optimal demonstrations \cite{ArgallChernovaVelosoBrowning09,Pomerleau89} 
while the second, called inverse RL (IRL) \cite{NgRussell00,russell1998learning}, tries to first recover a reward function that explains the demonstrations and then computes an optimal policy from it.
The direct methods based on supervised learning 
usually suffer when the reward function is sparse and even more when dynamics is also perturbed 
\cite{PiotGeistPietquin13}.

As the reward function is generally considered to be a more compact, robust and transferable representation of a task than a policy \cite{AbbeelNg04,russell1998learning}, we only discuss reward learning approaches here.

As for many inverse problems, IRL is ill-posed:
any constant function is a trivial solution that makes all policies equivalent and therefore optimal.
Various solutions were proposed to tackle this degeneracy issue, differing on whether a probabilistic model is assumed or not on the generation of the observation.
When the state and/or action spaces are large, the reward function is generally assumed to take a parametric form: $\Rewards(\state, \action) = f_{\bm\theta}(\state, \action)$ for $f_{\bm\theta}$ a parametric function of $\bm\theta$.
One important case, called {\em linear features}, is when $f$ is linear in $\bm\theta$, i.e., $\Rewards(\state, \action) = \sum_i \theta_i \phi_i(\state, \action)$ where $\phi_i$ are basis functions.

{\bf No generative model assumption.}
As underlined by \textcite{NeuSzepesvari09}, many IRL methods can be viewed as finding the reward function $\Rewards$ that minimizes a dissimilarity measure between the policy $\pi_\Rewards^*$ optimal for $\Rewards$ and the expert demonstrations.
Most work assume a linear-feature reward function, with some exceptions that we mention below.
\textcite{AbbeelNg04} introduced the important idea of expected feature matching, 
which says that the expected features of $\pi_\Rewards^*$ and those estimated from the demonstrations should be close.
Thus, they notably proposed the projection method, which amounts to minimizing the Euclidean distance between those two expected features.
\textcite{NeuSzepesvari07} proposed a natural gradient method for minimizing this objective function.
\textcite{SyedSchapire08} reformulated the projection method problem as a zero-sum two-player game, with the nice property that the learned policy may perform better than the demonstrated one. 
\textcite{AbbeelNg04}'s work was extended to the partially observable case \cite{ChoiKim11}.

Besides, \textcite{RatliffBagnellZinkevich06} proposed a max-margin approach enforcing that the found solution is better than any other one by at least a margin. 
Interestingly, the method can learn from multiple MDPs.
It was later extended to the non-linear feature case \cite{RatliffBradleyBagnellChestnutt07}.

Another technique \cite{KleinGeistPiotPietquin12,PiotGeistPietquin14} consists in learning a classifier based on a linearly parametrized score function to predict the best action for a state given the set of demonstrations.
The learned score function can then be interpreted as a value function and can be used to recover a reward function.

Traditional IRL methods learn from (near) optimal demonstration.
More recent approaches extend IRL to learn from other types of observations, e.g., a set of (non-necessarily optimal) demonstrations rated by an expert \cite{El-AsriPiotGeistLarochePietquin16,BurchfieldTomasiParr16}, bad demonstrations \cite{SebagAkrourMayeurSchoenauer16} or pairwise comparisons \cite{da-SilvaCostaLima06,WirthNeumann15}.
In the latter case, the interactive setting is investigated with a reliable expert \cite{ChernovaVeloso09} or unreliable one \cite{WengBusaFeketeHullermeier13}.

{\bf Generative model assumption.}
Another way to tackle the degeneracy issue is to assume a probabilistic model on how observations are generated.
Here, most work assumes that the expert policy is described by Boltzmann distributions, where higher-valued actions are more probable.
Two notable exceptions are the work of \textcite{GrollmanBillard11}, which shows how to learn from failed demonstration\PW{s} using Gaussian mixture models, and the Bayesian approach of \textcite{RamachandranAmir07}, with the assumption that state-action pairs in demonstrations follow such a Boltzmann distribution.
This latter approach has been extended to Boltzmann distribution-based expert policy and for multi-task learning \cite{DimitrakakisRothkopf11}, and to account for multiple reward functions \cite{ChoiKim12}.
This Bayesian approach has been investigated to interactive settings where the agent can query for an optimal demonstration in a chosen state \cite{LopesMeloMontesano09} or for a pairwise comparison  \cite{WilsonFernTadepalli12,AkrourSchoenauerSebag13,AkrourSchoenauerSoupletSebag14}.

Without assuming a prior, \textcite{Babes-VromanMarivateSubramanianLittman11} proposed to recover the expert reward function by maximum likelihood. 
The approach is able to handle the possibility of multiple intentions in the demonstrations.
Furthermore, \textcite{NguyenLowJaillet15} suggested an Expectation-Maximization approach to learn from demonstration induced by locally consistent reward functions.

To tackle the degeneracy issue, \textcite{ZiebartMaasBagnellDey10} argued for the use of the maximum entropy principle, which states that among all solutions that fit the observations, the least informative one (i.e., maximum entropy) should be chosen, with the assumption that a reward function induces a Boltzmann probability distribution over trajectories. 
When the transition function is not known, \textcite{BoulariasKoberPeters11} extended this approach by proposing to minimize the relative entropy between the probability distribution (over trajectories) induced by a policy and a baseline distribution under an expected feature matching constraint.	
\textcite{WulfmeierOndruskaPosner15} extended this approach to the case where a deep neural network is used for the representation of the reward function, while \textcite{BogertLinDoshiKulic16} took into account non-observable variables.

\subsection{Preference-Based Approaches} \label{sec:preferencebased}

Another line of work redefines policy optimality directly based on pairwise comparisons of histories without assuming the existence of a scalar numeric reward function.
This notably accounts for situations where reward values and probabilities are not commensurable.
In this context, different decision criteria (e.g., quantile \cite{GilbertWeng16}) may be used.
One popular decision model (\cite{YueBroderKleinbergJoachims12,FurnkranzHullermeierChengPark12}) is defined as follows: a policy $\pi$ is preferred to another policy $\pi'$ if
\begin{align}
\mathbb{P}[h^\pi \succsim h^{\pi'}] \ge \mathbb{P}[h^{\pi'} \succsim h^\pi], \label{eq:probabilisticdominance}
\end{align}
where $\succsim$ is a preorder over histories, $h^\pi$ is a random variable representing the history generated by policy $\pi$ and therefore $\mathbb{P}[h^\pi \succsim h^{\pi'}]$ is the probability that a history generated by $\pi$ is not less preferred than a history generated by $\pi'$.
Based on (\ref{eq:probabilisticdominance}), \textcite{FurnkranzHullermeierChengPark12} proposed a policy iteration algorithm.
However, one crucial issue with (\ref{eq:probabilisticdominance}) is that the concept of optimal solution is not well-defined as (\ref{eq:probabilisticdominance}) can lead to preference cycles \cite{GilbertSpanjaardViappianiWeng15}.
\textcite{BusaFeketeSzorenyiWengChengHullermeier14} circumvented this problem by refining this decision model with criteria from social choice theory.
In \cite{GilbertSpanjaardViappianiWeng15}, the issue was solved by considering mixed solutions: an optimal mixed solution is guaranteed to exist by interpreting it as a Nash equilibrium of a two-player zero-sum game.
\textcite{GilbertZanuttiniViappianiWengNicart16} proposed a model-free RL algorithm based on a two-timescale technique to find such a mixed optimal solution.

\subsection{Risk-Sensitive Criteria} \label{sec:risksensitive}

Taking into account risk is important in decision-making under uncertainty (see Chapter 17 of this volume). 
The standard criterion based on expectation is risk-neutral.
When it is known that a policy will only be used a few limited number of times, variability in the obtained rewards should be penalized.
Besides, in some hazardous domains, good policies need to absolutely avoid bad or error states. 
In those two cases, preferences over policies need to be defined to be risk-sensitive.

In its simplest form, risk can directly be represented as a probability.
For instance, \textcite{GeibelWysotzky05} adopted such an approach and consider MDP problems with two objectives where the first objective is the standard decision criterion and the second objective is to minimize the probability of reaching a set of bad states.

A more advanced approach is based on risk-sensitive decision criteria \cite{BarberaHammondSeidl99}.
Variants of Expected Utility \cite{Machina88}, which is the standard risk-sensitive criterion, were investigated in two cases when the utility function is exponential \cite{Borkar10,MoldovanAbbeel12} and when it is quadratic \cite{TamarDi-CastroMannor12,TamarDi-CastroMannor13,Gosavi14}. 
In the latter case, the criterion amounts to penalizing the standard criterion by the variance of the cumulated reward.
While the usual approach is to transform the cumulated reward, \textcite{MihatschNeuneier02} proposed to directly transform the temporal differences during learning.

Other approaches consider risk measures \cite{DenuitDhaeneGoovaertsKaasLaeven06} and in particular coherent risk measures \cite{ArtznerDelbaenEberHeath99}. 
Value-at-risk, popular in finance, was considered in \cite{GilbertWeng16}.
Policy gradient methods \cite{ChowGhavamzadeh14,TamarGlassnerMannor15} were proposed to optimize Conditional Value-at-Risk (CVaR) and were extended to any coherent risk measure \cite{TamarChowGhavamzadehMannor15}.
\textcite{JiangPowell16} proposed dynamic quantile-based risk measures, which encompasses VaR and CVaR, and investigated an approximate dynamic programming scheme to optimize them.

In risk-constrained problems, the goal is to maximize the expectation of return while bounding a risk measure.
For variance-constrained problems, \textcite{PrashanthGhavamzadeh16} proposed an actor-critic algorithm.
For CVaR-constrained problems, \textcite{BorkarJain14} proposed a two-timescale stochastic approximation technique, while \textcite{ChowGhavamzadehJansonPavone16} investigated policy gradient and actor-critic methods.

One important issue to consider when dealing with risk-sensitive criteria is that the Bellman optimality principle generally does not hold anymore:
a sub-policy of an optimal risk-sensitive policy may not be optimal.
However, in most cases, the Bellman optimality principle may be recovered by considering a state-augmented MDP, where the state includes the rewards cumulated so far \cite{LiuKoenig06}. 

\section{Conclusion}\label{sec:conclusion}

Recently, thanks to a number of success stories, reinforcement learning (RL) has become a very active research area. 
In this chapter, we recalled the basic setting of RL. 
Our focus was to present an overview of the main techniques, which can be divided into value-based and policy search methods, for solving large-sized RL problems with function approximation. 
We also presented some approaches for tackling the issue of unknown rewards that a system designer would encounter in practice and recalled some recent work in RL when risk-sensitivity needs to be taken into account in decision-making.

Currently RL still has too large sample and computational requirements for many practical domains (e.g., robotics).
Research work is very active on improving RL algorithms along those two dimensions, notably by exploiting the structure of the problem \cite{KulkarniNarasimhanSaeediTenenbaum16} or other a priori knowledge, expressed in temporal logic \cite{WenPapushaTopcu17} for instance, or by reusing previous learning experience with transfer learning \cite{TaylorStone09}, lifelong learning \cite{Bou-AmmarTutunovEaton15}, multi-task learning \cite{WilsonFernRayTadepalli07} or curriculum learning \cite{WuTian17}, to cite a few.
Having more efficient RL algorithms is important as it will pave the way to more applications in more realistic domains.

\bibliography{biblio-vol1-chapRL-PW,biblio-vol1-chapRL-OB,biblio-vol1-chapRL-OP}

\begin{thebibliography}{}

\bibitem[Abbeel et~al., 2010]{AbbeelCoatesNg10}
Abbeel, P., Coates, A., and Ng, A.~Y. (2010).
\newblock Autonomous helicopter aerobatics through apprenticeship learning.
\newblock {\em International Journal of Robotics Research}, 29(13):1608--1639.

\bibitem[Abbeel and Ng, 2004]{AbbeelNg04}
Abbeel, P. and Ng, A. (2004).
\newblock Apprenticeship learning via inverse reinforcement learning.
\newblock In {\em International Conference Machine Learning}.

\bibitem[Akrour et~al., 2013]{AkrourSchoenauerSebag13}
Akrour, R., Schoenauer, M., and Sebag, M. (2013).
\newblock Interactive robot education.
\newblock In {\em ECML PKDD, Lecture Notes in Computer Science}.

\bibitem[Akrour et~al., 2014]{AkrourSchoenauerSoupletSebag14}
Akrour, R., Schoenauer, M., Souplet, J.-C., and Sebag, M. (2014).
\newblock Programming by feedback.
\newblock In {\em ICML}.

\bibitem[Anderson and Moore, 2005]{AndMoo-of05}
Anderson, B. D.~O. and Moore, J.~B. (2005).
\newblock {\em Optimal Filtering}.
\newblock Dover Publications.

\bibitem[Antos et~al., 2008]{antos2008fitted}
Antos, A., Szepesv{\'a}ri, C., and Munos, R. (2008).
\newblock Fitted q-iteration in continuous action-space mdps.
\newblock In {\em Advances in neural information processing systems}, pages
  9--16.

\bibitem[Argall et~al., 2009]{ArgallChernovaVelosoBrowning09}
Argall, B., Chernova, S., Veloso, M., and Browning, B. (2009).
\newblock A survey of robot learning from demonstration.
\newblock {\em Robotics and Autonomous Systems}, 57(5):469--483.

\bibitem[Artzner et~al., 1999]{ArtznerDelbaenEberHeath99}
Artzner, P., Delbaen, F., Eber, J., and Heath, D. (1999).
\newblock Coherent measures of risk.
\newblock {\em Mathematical Finance}, 9(3):203--228.

\bibitem[Babes-Vroman et~al., 2011]{Babes-VromanMarivateSubramanianLittman11}
Babes-Vroman, M., Marivate, V., Subramanian, K., and Littman, M. (2011).
\newblock Apprenticeship learning about multiple intentions.
\newblock In {\em ICML}.

\bibitem[Bagnell and Schneider, 2001]{BagSch-icra01}
Bagnell, J.~A. and Schneider, J.~G. (2001).
\newblock Autonomous helicopter control using reinforcement learning policy
  search methods.
\newblock In {\em Proceedings of the International Conference on Robotics and
  Automation}, pages 1615--1620.

\bibitem[Bagnell and Schneider, 2003]{BagSch-ijcai03}
Bagnell, J.~A. and Schneider, J.~G. (2003).
\newblock Covariant policy search.
\newblock In {\em Proceedings of the International Joint Conference on
  Artifical Intelligence}.

\bibitem[Bai et~al., 2013]{BaiWuChen13}
Bai, A., Wu, F., and Chen, X. (2013).
\newblock Towards a principled solution to simulated robot soccer.
\newblock In {\em RoboCup-2012: Robot Soccer World Cup XVI, Lecture Notes in
  Artificial Intelligence}, volume 7500.

\bibitem[Baird et~al., 1995]{baird1995residual}
Baird, L. et~al. (1995).
\newblock Residual algorithms: Reinforcement learning with function
  approximation.
\newblock In {\em Proceedings of the twelfth international conference on
  machine learning}, pages 30--37.

\bibitem[Barbera et~al., 1999]{BarberaHammondSeidl99}
Barbera, S., Hammond, P., and Seidl, C. (1999).
\newblock {\em Handbook of Utility Theory}.
\newblock Springer.

\bibitem[B{\"a}uerle and Rieder, 2011]{BauerleRieder11}
B{\"a}uerle, N. and Rieder, U. (2011).
\newblock {\em Markov Decision Processes with Applications to Finance}.
\newblock Springer Science \& Business Media.

\bibitem[Baxter and Bartlett, 2001]{BaxBar-jmlr01}
Baxter, J. and Bartlett, P. (2001).
\newblock Infinite-horizon policy-gradient estimation.
\newblock {\em Journal of Artificial Intelligence Research}, 15:319--350.

\bibitem[Baxter et~al., 2001]{BaxBarWea-jmlr01}
Baxter, J., Bartlett, P., and Weaver, L. (2001).
\newblock Experiments with infinite-horizon, policy-gradient estimation.
\newblock {\em Journal of Artificial Intelligence Research}, 15:351--381.

\bibitem[Bellman and Dreyfus, 1959]{BellmanPoly}
Bellman, R. and Dreyfus, S. (1959).
\newblock Functional approximations and dynamic programming.
\newblock {\em Mathematical Tables and Other Aids to Computation},
  13(68):247--251.

\bibitem[Bellman et~al., 1963]{bellman1963polynomial}
Bellman, R., Kalaba, R., and Kotkin, B. (1963).
\newblock Polynomial approximation--a new computational technique in dynamic
  programming: Allocation processes.
\newblock {\em Mathematics of Computation}, 17(82):155--161.

\bibitem[Bogert et~al., 2016]{BogertLinDoshiKulic16}
Bogert, K., Lin, J. F.-S., Doshi, P., and Kulic, D. (2016).
\newblock Expectation-maximization for inverse reinforcement learning with
  hidden data.
\newblock In {\em AAMAS}.

\bibitem[Bojarski et~al.,
  2016]{BojarskiTestaDworakowskiFirnerFleppGoyalJackelMonfortMullerZhangZhangZhao16}
Bojarski, M., Testa, D.~D., Dworakowski, D., Firner, B., Flepp, B., Goyal, P.,
  Jackel, L.~D., Monfort, M., Muller, U., Zhang, J., Zhang, X., and Zhao, J.
  (2016).
\newblock End to end learning for self-driving cars.
\newblock Technical report, NVIDIA.

\bibitem[Borkar and Jain, 2014]{BorkarJain14}
Borkar, V. and Jain, R. (2014).
\newblock Risk-constrained {M}arkov decision processes.
\newblock {\em IEEE Transactions on Automatic Control}, 59(9):2574--2579.

\bibitem[Borkar, 2010]{Borkar10}
Borkar, V.~S. (2010).
\newblock Learning algorithms for risk-sensitive control.
\newblock In {\em International Symposium on Mathematical Theory of Networks
  and Systems}.

\bibitem[{Bou Ammar} et~al., 2015]{Bou-AmmarTutunovEaton15}
{Bou Ammar}, H., Tutunov, R., and Eaton, E. (2015).
\newblock Safe policy search for lifelong reinforcement learning with sublinear
  regret.
\newblock In {\em ICML}.

\bibitem[Boularias et~al., 2011]{BoulariasKoberPeters11}
Boularias, A., Kober, J., and Peters, J. (2011).
\newblock Relative entropy inverse reinforcement learning.
\newblock In {\em AISTATS}.

\bibitem[Boutilier et~al., 1995]{BouDeaGol-ijcai95}
Boutilier, C., Dearden, R., and Goldszmidt, M. (1995).
\newblock Exploiting structure in policy construction.
\newblock In {\em Proceedings of the Fourteenth International Joint Conference
  on Artificial Intelligence}, pages 1104--1111.

\bibitem[Boutilier et~al., 2000]{BoutilierDeardenGoldszmidt00}
Boutilier, C., Dearden, R., and Goldszmidt, M. (2000).
\newblock Stochastic dynamic programming with factored representations.
\newblock {\em Artificial Intelligence}, 121(1-2):49--107.

\bibitem[Bradtke and Barto, 1996]{bradtke1996linear}
Bradtke, S.~J. and Barto, A.~G. (1996).
\newblock Linear least-squares algorithms for temporal difference learning.
\newblock {\em Machine Learning}, 22:33--57.

\bibitem[Burchfield et~al., 2016]{BurchfieldTomasiParr16}
Burchfield, B., Tomasi, C., and Parr, R. (2016).
\newblock Distance minimization for reward learning from scored trajectories.
\newblock In {\em AAAI}.

\bibitem[Busa-Fekete et~al., 2014]{BusaFeketeSzorenyiWengChengHullermeier14}
Busa-Fekete, R., Sz{\"{o}}renyi, B., Weng, P., Cheng, W., and
  H{\"{u}}llermeier, E. (2014).
\newblock {P}reference-based {R}einforcement {L}earning: {E}volutionary
  {D}irect {P}olicy {S}earch using a {P}reference-based {R}acing {A}lgorithm.
\newblock {\em {M}achine {L}earning}, 97(3):327--351.

\bibitem[Busoniu et~al., 2010]{BusoniuBabuskaDeSchutter10}
Busoniu, L., Babuska, R., and {De Schutter}, B. (2010).
\newblock {\em Innovations in Multi-Agent Systems and Applications -- 1},
  volume 310, chapter "Multi-agent reinforcement learning: An overview,"
  Chapter 7, pages 183--221.
\newblock {S}pringer.

\bibitem[Chernova and Veloso, 2009]{ChernovaVeloso09}
Chernova, S. and Veloso, M. (2009).
\newblock Interactive policy learning through confidence-based autonomy.
\newblock {\em Journal of Artificial Intelligence Research}, 34:1--25.

\bibitem[Choi and Van~Roy, 2006]{choi2006generalized}
Choi, D. and Van~Roy, B. (2006).
\newblock A generalized kalman filter for fixed point approximation and
  efficient temporal-difference learning.
\newblock {\em Discrete Event Dynamic Systems}, 16(2):207--239.

\bibitem[Choi and Kim, 2011]{ChoiKim11}
Choi, J. and Kim, K.-E. (2011).
\newblock Inverse reinforcement learning in partially observable environments.
\newblock {\em JMLR}, 12:691--730.

\bibitem[Choi and Kim, 2012]{ChoiKim12}
Choi, J. and Kim, K.-E. (2012).
\newblock Nonparametric bayesian inverse reinforcement learning for multiple
  reward functions.
\newblock In {\em NIPS}.

\bibitem[Chow and Ghavamzadeh, 2014]{ChowGhavamzadeh14}
Chow, Y. and Ghavamzadeh, M. (2014).
\newblock Algorithms for cvar optimization in {MDPs}.
\newblock In {\em NIPS}.

\bibitem[Chow et~al., 2017]{ChowGhavamzadehJansonPavone16}
Chow, Y., Ghavamzadeh, M., Janson, L., and Pavone, M. (2017).
\newblock Risk-constrained reinforcement learning with percentile risk
  criteria.
\newblock {\em JMLR}, 18(1).

\bibitem[{da Silva} et~al., 2006]{da-SilvaCostaLima06}
{da Silva}, V.~F., Costa, A. H.~R., and Lima, P. (2006).
\newblock Inverse reinforcement learning with evaluation.
\newblock In {\em IEEE ICRA}.

\bibitem[Daniel et~al., 2012]{DanNeuPet-aistats12}
Daniel, C., Neumann, G., and Peters, J. (2012).
\newblock Hierarchical relative entropy policy search.
\newblock In {\em Proceedings of the International Conference of Artificial
  Intelligence and Statistics}, pages 273--281.

\bibitem[de~Boer et~al., 2005]{BoeKroManRub-aor05}
de~Boer, P., Kroese, D., Mannor, S., and Rubinstein, R. (2005).
\newblock A tutorial on the cross-entropy method.
\newblock {\em Annals of Operations Research}, 134(1):19--67.

\bibitem[{de Farias} and {Van Roy}, 2003]{de-FariasVan-Roy03}
{de Farias}, D. and {Van Roy}, B. (2003).
\newblock The linear programming approach to approximate dynamic programming.
\newblock {\em Operations Research}, 51(6):850--865.

\bibitem[Degris et~al., 2006]{DegSigWui-icml06}
Degris, T., Sigaud, O., and Wuillemin, P.-H. (2006).
\newblock Learning the structure of {Factored Markov Decision Processes} in
  reinforcement learning problems.
\newblock In {\em Proceedings of the 23rd International Conference on Machine
  Learning}.

\bibitem[Deisenroth et~al., 2011]{DeiNeuPet-ftr11}
Deisenroth, M.~P., Neumann, G., and Peters, J. (2011).
\newblock A survey on policy search for robotics.
\newblock {\em Foundations and Trends in Robotics}, 2(1--2):1--142.

\bibitem[Deisenroth and Rasmussen, 2011]{DeiRas-icml11}
Deisenroth, M.~P. and Rasmussen, C.~E. (2011).
\newblock {PILCO}: A model-based and data-efficient approach to policy search.
\newblock In {\em Proceedings of the International Conference on Machine
  Learning}, pages 465--472.

\bibitem[Denuit et~al., 2006]{DenuitDhaeneGoovaertsKaasLaeven06}
Denuit, M., Dhaene, J., Goovaerts, M., Kaas, R., and Laeven, R. (2006).
\newblock Risk measurement with equivalent utility principles.
\newblock {\em Statistics and Decisions}, 24:1--25.

\bibitem[Dimitrakakis and Rothkopf, 2011]{DimitrakakisRothkopf11}
Dimitrakakis, C. and Rothkopf, C.~A. (2011).
\newblock Bayesian multitask inverse reinforcement learning.
\newblock In {\em EWRL}.

\bibitem[{El Asri} et~al., 2012]{El-AsriLarochePietquin12}
{El Asri}, L., Laroche, R., and Pietquin, O. (2012).
\newblock Reward function learning for dialogue management.
\newblock In {\em STAIRS}.

\bibitem[{El Asri} et~al., 2016]{El-AsriPiotGeistLarochePietquin16}
{El Asri}, L., Piot, B., Geist, M., Laroche, R., and Pietquin, O. (2016).
\newblock Score-based inverse reinforcement learning.
\newblock In {\em AAMAS}.

\bibitem[Engel et~al., 2005]{GPTD}
Engel, Y., Mannor, S., and Meir, R. (2005).
\newblock Reinforcement learning with gaussian processes.
\newblock In {\em Proceedings of the 22nd international conference on Machine
  learning}, pages 201--208. ACM.

\bibitem[Ernst et~al., 2005]{ernst2005tree}
Ernst, D., Geurts, P., and Wehenkel, L. (2005).
\newblock Tree-based batch mode reinforcement learning.
\newblock {\em Journal of Machine Learning Research}, 6(Apr):503--556.

\bibitem[F\"urnkranz et~al., 2012]{FurnkranzHullermeierChengPark12}
F\"urnkranz, J., H\"ullermeier, E., Cheng, W., and Park, S. (2012).
\newblock Preference-based reinforcement learning: A formal framework and a
  policy iteration algorithm.
\newblock {\em Machine Learning}, 89(1):123--156.

\bibitem[Geibel and Wysotzky, 2005]{GeibelWysotzky05}
Geibel, P. and Wysotzky, F. (2005).
\newblock Risk-sensitive reinforcement learning applied to control under
  constraints.
\newblock {\em JAIR}, 24:81--108.

\bibitem[Geist and Pietquin, 2010a]{KTD}
Geist, M. and Pietquin, O. (2010a).
\newblock Kalman temporal differences.
\newblock {\em Journal of artificial intelligence research}, 39:483--532.

\bibitem[Geist and Pietquin, 2010b]{geist2010statistically}
Geist, M. and Pietquin, O. (2010b).
\newblock Statistically linearized least-squares temporal differences.
\newblock In {\em Ultra Modern Telecommunications and Control Systems and
  Workshops (ICUMT), 2010 International Congress on}, pages 450--457. IEEE.

\bibitem[Geist and Pietquin, 2011]{GeistPietquin11}
Geist, M. and Pietquin, O. (2011).
\newblock Parametric value function approximation: a unified view.
\newblock In {\em ADPRL}.

\bibitem[Geist and Pietquin, 2013]{geist2013algorithmic}
Geist, M. and Pietquin, O. (2013).
\newblock Algorithmic survey of parametric value function approximation.
\newblock {\em IEEE Transactions on Neural Networks and Learning Systems},
  24(6):845--867.

\bibitem[Ghavamzadeh et~al., 2015]{GhavamzadehMannorPineauTamar15}
Ghavamzadeh, M., Mannor, S., Pineau, J., and Tamar, A. (2015).
\newblock Bayesian reinforcement learning: a survey.
\newblock {\em Foundations and Trends in Machine Learning}, 8(5--6):359--492.

\bibitem[Gilbert et~al., 2015]{GilbertSpanjaardViappianiWeng15}
Gilbert, H., Spanjaard, O., Viappiani, P., and Weng, P. (2015).
\newblock Solving {MDP}s with skew symmetric bilinear utility functions.
\newblock In {\em {International Joint Conference in Artificial Intelligence
  (IJCAI)}}, pages 1989--1995.

\bibitem[Gilbert and Weng, 2016]{GilbertWeng16}
Gilbert, H. and Weng, P. (2016).
\newblock Quantile reinforcement learning.
\newblock In {\em Asian Workshop on Reinforcement Learning}.

\bibitem[Gilbert et~al., 2016]{GilbertZanuttiniViappianiWengNicart16}
Gilbert, H., Zanuttini, B., Viappiani, P., Weng, P., and Nicart, E. (2016).
\newblock Model-free reinforcement learning with skew-symmetric bilinear
  utilities.
\newblock In {\em {International Conference on Uncertainty in Artificial
  Intelligence (UAI)}}.

\bibitem[Gordon, 1995]{gordon1995stable}
Gordon, G.~J. (1995).
\newblock Stable function approximation in dynamic programming.
\newblock In {\em Proceedings of the twelfth international conference on
  machine learning}, pages 261--268.

\bibitem[Gosavi, 2014]{Gosavi14}
Gosavi, A.~A. (2014).
\newblock Variance-penalized markov decision processes: Dynamic programming and
  reinforcement learning techniques.
\newblock {\em International Journal of General Systems}, 43(6):649--669.

\bibitem[Grollman and Billard, 2011]{GrollmanBillard11}
Grollman, D.~H. and Billard, A. (2011).
\newblock Donut as i do: learning from failed demonstrations.
\newblock In {\em IEEE ICRA}.

\bibitem[Guestrin et~al., 2004]{GuestrinHauskrechtKveton04}
Guestrin, C., Hauskrecht, M., and Kveton, B. (2004).
\newblock Solving factored {MDPs} with continuous and discrete variables.
\newblock In {\em AAAI}, pages 235--242.

\bibitem[Hansen et~al., 2003]{HanMulKou-ec03}
Hansen, N., Muller, S., and Koumoutsakos, P. (2003).
\newblock Reducing the time complexity of the derandomized evolution strategy
  with covariance matrix adaptation {(CMA-ES)}.
\newblock {\em Evolutionary Computation}, 11(1):1--18.

\bibitem[Heidrich-Meisner and Igel, 2009]{HeiIge-joa09}
Heidrich-Meisner, V. and Igel, C. (2009).
\newblock Neuroevolution strategies for episodic reinforcement learning.
\newblock {\em Journal of Algorithms}, 64(4):152--168.

\bibitem[Hussein et~al., 2017]{HusseinGaberElyanJayne17}
Hussein, A., Gaber, M.~M., Elyan, E., and Jayne, C. (2017).
\newblock Imitation learning: a survey of learning methods.
\newblock {\em ACM Computing Surveys}.

\bibitem[Jiang and Powell, 2018]{JiangPowell16}
Jiang, D.~R. and Powell, W.~B. (2018).
\newblock Risk-averse approximate dynamic programming with quantile-based risk
  measures.
\newblock {\em Mathematics of Operations Research}, 43(2):347--692.

\bibitem[Julier and Uhlmann, 2004]{JulUhl-ieee04}
Julier, S.~J. and Uhlmann, J.~K. (2004).
\newblock Unscented filtering and nonlinear estimation.
\newblock {\em Proceedings of the {IEEE}}, 92(3):401--422.

\bibitem[Klein et~al., 2012]{KleinGeistPiotPietquin12}
Klein, E., Geist, M., Piot, B., and Pietquin, O. (2012).
\newblock Inverse reinforcement learning through structured classification.
\newblock In {\em NIPS}.

\bibitem[Kober et~al., 2010]{KobOztPet-rss10}
Kober, J., Oztop, E., and Peters, J. (2010).
\newblock Reinforcement learning to adjust robot movements to new situations.
\newblock In {\em Proceedings of the 2010 Robotics: Science and Systems
  Conference}.

\bibitem[Kober and Peters, 2010]{KobPet-ml10}
Kober, J. and Peters, J. (2010).
\newblock Policy search for motor primitives in robotics.
\newblock {\em Machine Learning}, pages 1--33.

\bibitem[Kulkarni et~al., 2016]{KulkarniNarasimhanSaeediTenenbaum16}
Kulkarni, T., Narasimhan, K.~R., Saeedi, A., and Tenenbaum, J. (2016).
\newblock Hierarchical deep reinforcement learning: Integrating temporal
  abstraction and intrinsic motivation.
\newblock In {\em NIPS}.

\bibitem[Lagoudakis and Parr, 2003]{lagoudakis2003least}
Lagoudakis, M.~G. and Parr, R. (2003).
\newblock Least-squares policy iteration.
\newblock {\em Journal of machine learning research}, 4(Dec):1107--1149.

\bibitem[LeCun et~al., 2015]{lecun2015deep}
LeCun, Y., Bengio, Y., and Hinton, G. (2015).
\newblock Deep learning.
\newblock {\em Nature}, 521(7553):436--444.

\bibitem[Lesner and Zanuttini, 2011]{LesZan-ewrl11}
Lesner, B. and Zanuttini, B. (2011).
\newblock Handling ambiguous effects in action learning.
\newblock In {\em Proceedings of the 9th European Workshop on Reinforcement
  Learning}, page~12.

\bibitem[Lillicrap et~al.,
  2016]{LillicrapHuntPritzelHeessErezTassaSilverWierstra16}
Lillicrap, T.~P., Hunt, J.~J., Pritzel, A., Heess, N., Erez, T., Tassa, Y.,
  Silver, D., and Wierstra, D. (2016).
\newblock Continuous control with deep reinforcement learning.
\newblock In {\em ICLR}.

\bibitem[Lin, 1992]{lin1992self}
Lin, L.-H. (1992).
\newblock Self-improving reactive agents based on reinforcement learning,
  planning and teaching.
\newblock {\em Machine learning}, 8(3/4):69--97.

\bibitem[Liu and Koenig, 2006]{LiuKoenig06}
Liu, Y. and Koenig, S. (2006).
\newblock Functional value iteration for decision-theoretic planning with
  general utility functions.
\newblock In {\em AAAI}, pages 1186--1193. AAAI.

\bibitem[Lopes et~al., 2009]{LopesMeloMontesano09}
Lopes, M., Melo, F., and Montesano, L. (2009).
\newblock Active learning for reward estimation in inverse reinforcement
  learning.
\newblock In {\em ECML/PKDD, Lecture Notes in Computer Science}, volume 5782,
  pages 31--46.

\bibitem[Machina, 1988]{Machina88}
Machina, M. (1988).
\newblock Expected utility hypothesis.
\newblock In Eatwell, J., Milgate, M., and Newman, P., editors, {\em The New
  Palgrave: A Dictionary of Economics}, pages 232--239. Macmillan.

\bibitem[Matignon et~al., 2006]{MatignonLaurentLe-Fort-Piat06}
Matignon, L., Laurent, G.~J., and {Le Fort-Piat}, N. (2006).
\newblock Reward function and initial values: Better choices for accelerated
  goal-directed reinforcement learning.
\newblock {\em Lecture notes in CS}, 1(4131):840--849.

\bibitem[Mihatsch and Neuneier, 2002]{MihatschNeuneier02}
Mihatsch, O. and Neuneier, R. (2002).
\newblock Risk-sensitive reinforcement learning.
\newblock {\em {M}achine {L}earning}.

\bibitem[Mnih et~al.,
  2016]{MnihBadiaMirzaGravesLillicrapHarleySilverKavukcuoglu16}
Mnih, V., Badia, A.~P., Mirza, M., Graves, A., Lillicrap, T.~P., Harley, T.,
  Silver, D., and Kavukcuoglu, K. (2016).
\newblock Asynchronous methods for deep reinforcement learning.
\newblock In {\em ICML}.

\bibitem[Mnih et~al.,
  2015]{MnihKavukcuogluSilverRusuVenessBellemareGravesRiedmillerFidjelandOstrovskiPetersenBeattieSadikAntonoglouKingKumaranWierstraLeggHassabis15}
Mnih, V., Kavukcuoglu, K., Silver, D., Rusu, A.~A., Veness, J., Bellemare,
  M.~G., Graves, A., Riedmiller, M., Fidjeland, A.~K., Ostrovski, G., Petersen,
  S., Beattie, C., Sadik, A., Antonoglou, I., King, H., Kumaran, D., Wierstra,
  D., Legg, S., and Hassabis, D. (2015).
\newblock Human-level control through deep reinforcement learning.
\newblock {\em Nature}, 518:529--533.

\bibitem[Moldovan and Abbeel, 2012]{MoldovanAbbeel12}
Moldovan, T. and Abbeel, P. (2012).
\newblock Risk aversion {M}arkov decision processes via near-optimal {C}hernoff
  bounds.
\newblock In {\em NIPS}.

\bibitem[Neu and Szepesvari, 2007]{NeuSzepesvari07}
Neu, G. and Szepesvari, C. (2007).
\newblock Apprenticeship learning using inverse reinforcement learning and
  gradient methods.
\newblock In {\em UAI}.

\bibitem[Neu and Szepesvari, 2009]{NeuSzepesvari09}
Neu, G. and Szepesvari, C. (2009).
\newblock Training parsers by inverse reinforcement learning.
\newblock {\em {M}achine {L}earning}, 77:303--337.

\bibitem[Neumann, 2011]{Neumann-icml11}
Neumann, G. (2011).
\newblock Variational inference for policy search in changing situations.
\newblock In {\em Proceedings of the International Conference on Machine
  Learning}, pages 817--824.

\bibitem[Ng and Russell, 2000]{NgRussell00}
Ng, A. and Russell, S. (2000).
\newblock Algorithms for inverse reinforcement learning.
\newblock In {\em ICML}. Morgan Kaufmann.

\bibitem[Ng and Jordan, 2000]{NgJor-uai00}
Ng, A.~Y. and Jordan, M. (2000).
\newblock {PEGASUS} : A policy search method for large {MDPs} and {POMDPs}.
\newblock In {\em Proceedings of the Conference on Uncertainty in Artificial
  Intelligence}.

\bibitem[Nguyen et~al., 2015]{NguyenLowJaillet15}
Nguyen, Q.~P., Low, K.~H., and Jaillet, P. (2015).
\newblock Inverse reinforcement learning with locally consistent reward
  functions.
\newblock In {\em NIPS}.

\bibitem[Pasula et~al., 2007]{PasZetKae-jair07}
Pasula, H.~M., Zettlemoyer, L.~S., and Kaelbling, L.~P. (2007).
\newblock Learning symbolic models of stochastic domains.
\newblock {\em Journal of Artificial Intelligence Research}, 29:309--352.

\bibitem[Peters et~al., 2010]{PetMueAlt-aaai10}
Peters, J., Mülling, K., and Altun, Y. (2010).
\newblock Relative entropy policy search.
\newblock In {\em Proceedings of the National Conference on Artificial
  Intelligence}.

\bibitem[Peters and Schaal, 2007]{PetSch-esann07}
Peters, J. and Schaal, S. (2007).
\newblock Applying the episodic natural actor-critic architecture to motor
  primitive learning.
\newblock In {\em Proceedings of the European Symposium on Artificial Neural
  Networks}.

\bibitem[Peters and Schaal, 2008a]{PetSch-nc08}
Peters, J. and Schaal, S. (2008a).
\newblock Natural actor-critic.
\newblock {\em Neurocomputation}, 71(7--9):1180--1190.

\bibitem[Peters and Schaal, 2008b]{PetSch-nn08}
Peters, J. and Schaal, S. (2008b).
\newblock Reinforcement learning of motor skills with policy gradients.
\newblock {\em Neural Networks}, 4:682--697.

\bibitem[Piot et~al., 2013]{PiotGeistPietquin13}
Piot, B., Geist, M., and Pietquin, O. (2013).
\newblock Learning from demonstrations: Is it worth estimating a reward
  function?
\newblock In {\em ECML PKDD, Lecture Notes in Computer Science}.

\bibitem[Piot et~al., 2014]{PiotGeistPietquin14}
Piot, B., Geist, M., and Pietquin, O. (2014).
\newblock {Boosted and Reward-regularized Classification for Apprenticeship
  Learning}.
\newblock In {\em AAMAS}, pages 1249--1256, Paris, France.

\bibitem[Pomerleau, 1989]{Pomerleau89}
Pomerleau, D. (1989).
\newblock Alvinn: An autonomous land vehicle in a neural network.
\newblock In {\em NIPS}.

\bibitem[Prashanth and Ghavamzadeh, 2016]{PrashanthGhavamzadeh16}
Prashanth, L. and Ghavamzadeh, M. (2016).
\newblock Variance-constrained actor-critic algorithms for discounted and
  average reward mdps.
\newblock {\em {M}achine {L}earning}.

\bibitem[Puterman, 1994]{Puterman94}
Puterman, M. (1994).
\newblock {\em Markov decision processes: discrete stochastic dynamic
  programming}.
\newblock Wiley.

\bibitem[Ramachandran and Amir, 2007]{RamachandranAmir07}
Ramachandran, D. and Amir, E. (2007).
\newblock Bayesian inverse reinforcement learning.
\newblock In {\em IJCAI}.

\bibitem[Randl\o{}v and Alstr\o{}m, 1998]{RandlovAlstrom98}
Randl\o{}v, J. and Alstr\o{}m, P. (1998).
\newblock Learning to drive a bicycle using reinforcement learning and shaping.
\newblock In {\em ICML, (1998).}

\bibitem[Ratliff et~al., 2006]{RatliffBagnellZinkevich06}
Ratliff, N., Bagnell, J., and Zinkevich, M. (2006).
\newblock Maximum margin planning.
\newblock In {\em ICML}.

\bibitem[Ratliff et~al., 2007]{RatliffBradleyBagnellChestnutt07}
Ratliff, N., Bradley, D., Bagnell, J.~A., and Chestnutt, J. (2007).
\newblock Boosting structured prediction for imitation learning.
\newblock In {\em NIPS}.

\bibitem[Riedmiller, 2005]{riedmiller2005neural}
Riedmiller, M. (2005).
\newblock Neural fitted q iteration-first experiences with a data efficient
  neural reinforcement learning method.
\newblock In {\em ECML}, volume 3720, pages 317--328. Springer.

\bibitem[Roijers et~al., 2013]{RoijersVamplewWhitesonDazeley13}
Roijers, D., Vamplew, P., Whiteson, S., and Dazeley, R. (2013).
\newblock A survey of multi-objective sequential decision-making.
\newblock {\em Journal of Artificial Intelligence Research}, 48:67--113.

\bibitem[Russell, 1998]{russell1998learning}
Russell, S. (1998).
\newblock Learning agents for uncertain environments.
\newblock In {\em Proceedings of the eleventh annual conference on
  Computational learning theory}, pages 101--103. ACM.

\bibitem[Samuel, 1959]{samuel1959some}
Samuel, A. (1959).
\newblock Some studies in machine learning using the game of checkers.
\newblock {\em IBM Journal of Research and Development}, 3(3):210--229.

\bibitem[Schaul et~al., 2016]{schaul2016prioritized}
Schaul, T., Quan, J., Antonoglou, I., and Silver, D. (2016).
\newblock Prioritized experience replay.
\newblock In {\em ICLR}.

\bibitem[Schulman et~al., 2015]{SchulmanLevineAbbeelJordanMoritz15}
Schulman, J., Levine, S., Abbeel, P., Jordan, M., and Moritz, P. (2015).
\newblock Trust region policy optimization.
\newblock In {\em ICML}.

\bibitem[Sebag et~al., 2016]{SebagAkrourMayeurSchoenauer16}
Sebag, M., Akrour, R., Mayeur, B., and Schoenauer, M. (2016).
\newblock Anti imitation-based policy learning.
\newblock In {\em ECML PKDD, Lecture Notes in Computer Science}.

\bibitem[Sehnke et~al., 2010]{SehEtAl-nn10}
Sehnke, F., Osendorfer, C., Rückstieß, T., Graves, A., Peters, J., and
  Schmidhuber, J. (2010).
\newblock Parameter-exploring policy gradients.
\newblock {\em Neural Networks}, 23(4):551--559.

\bibitem[Silver et~al.,
  2016]{SilverHuangMaddisonGuezSifreDriesscheSchrittwieserAntonoglouPanneerschelvamLanctotDielemanGreweNhamKalchbrennerSutskeverLillicrapLeachKavukcuogluGraepelHassabis16}
Silver, D., Huang, A., Maddison, C.~J., Guez, A., Sifre, L., {van den
  Driessche}, G., Schrittwieser, J., Antonoglou, I., Panneerschelvam, V.,
  Lanctot, M., Dieleman, S., Grewe, D., Nham, J., Kalchbrenner, N., Sutskever,
  I., Lillicrap, T., Leach, M., Kavukcuoglu, K., Graepel, T., and Hassabis, D.
  (2016).
\newblock Mastering the game of {Go} with deep neural networks and tree search.
\newblock {\em Nature}.

\bibitem[Silver et~al., 2014]{SilverLeverHeessDegrisWierstraRiedmiller14}
Silver, D., Lever, G., Heess, N., Degris, T., Wierstra, D., and Riedmiller, M.
  (2014).
\newblock Deterministic policy gradient algorithms.
\newblock In {\em ICML}.

\bibitem[Singh et~al., 1999]{SinghKearnsLitmanWalker99}
Singh, S., Kearns, M., Litman, D., and Walker, M. (1999).
\newblock Reinforcement learning for spoken dialogue systems.
\newblock In {\em NIPS}.

\bibitem[Spaan, 2012]{Spaan12}
Spaan, M.~T. (2012).
\newblock {\em Reinforcement Learning}, chapter Partially Observable Markov
  Decision Processes.
\newblock {S}pringer.

\bibitem[Sutton et~al.,
  2009]{SuttonMaeiPrecupBhatnagarSilverSzepesvariWiewiora09}
Sutton, R., Maei, H., Precup, D., Bhatnagar, S., Silver, D., Szepesv\'ari, C.,
  and Wiewiora, E. (2009).
\newblock Fast gradient-descent methods for temporal-difference learning with
  linear function approximation.
\newblock In {\em ICML}.

\bibitem[Syed and Schapire, 2008]{SyedSchapire08}
Syed, U. and Schapire, R.~E. (2008).
\newblock A game-theoretic approach to apprenticeship learning.
\newblock In {\em NIPS}.

\bibitem[Szita and L\"{o}rincz, 2006]{SziLor-nc05}
Szita, I. and L\"{o}rincz, A. (2006).
\newblock Learning tetris using the noisy cross-entropy method.
\newblock {\em Neural Computation}, 18:2936--2941.

\bibitem[Tamar et~al., 2015a]{TamarChowGhavamzadehMannor15}
Tamar, A., Chow, Y., Ghavamzadeh, M., and Mannor, S. (2015a).
\newblock Policy gradient for coherent risk measures.
\newblock In {\em NIPS}.

\bibitem[Tamar et~al., 2012]{TamarDi-CastroMannor12}
Tamar, A., {Di Castro}, D., and Mannor, S. (2012).
\newblock Policy gradient with variance related risk criteria.
\newblock In {\em ICML}.

\bibitem[Tamar et~al., 2013]{TamarDi-CastroMannor13}
Tamar, A., {Di Castro}, D., and Mannor, S. (2013).
\newblock Temporal difference methods for the variance of the reward to go.
\newblock In {\em ICML}.

\bibitem[Tamar et~al., 2015b]{TamarGlassnerMannor15}
Tamar, A., Glassner, Y., and Mannor, S. (2015b).
\newblock Optimizing the {CVaR} via sampling.
\newblock In {\em AAAI}.

\bibitem[Taylor and Stone, 2009]{TaylorStone09}
Taylor, M.~E. and Stone, P. (2009).
\newblock Transfer learning for reinforcement learning domains: A survey.
\newblock {\em Journal of Machine Learning Research}, 10:1633--1685.

\bibitem[Tesauro, 1995]{Tesauro95}
Tesauro, G. (1995).
\newblock Temporal difference learning and td-gammon.
\newblock {\em Communications of the ACM}, 38(3):58--68.

\bibitem[Van~Hasselt et~al., 2016]{van2016deep}
Van~Hasselt, H., Guez, A., and Silver, D. (2016).
\newblock Deep reinforcement learning with double q-learning.
\newblock In {\em AAAI}, pages 2094--2100.

\bibitem[van Otterlo, 2009]{Otterlo09}
van Otterlo, M. (2009).
\newblock {\em The Logic of Adaptive Behavior}.
\newblock IOS Press.

\bibitem[Walsh et~al., 2009]{WalSziDiuLit-uai09}
Walsh, T., Szita, I., Diuk, C., and Littman, M. (2009).
\newblock Exploring compact reinforcement-learning representations with linear
  regression.
\newblock In {\em Proceedings of the 25th Conference on Uncertainty in
  Artificial Intelligence}.

\bibitem[Wen et~al., 2017]{WenPapushaTopcu17}
Wen, M., Papusha, I., and Topcu, U. (2017).
\newblock Learning from demonstrations with high-level side information.
\newblock In {\em IJCAI}.

\bibitem[Weng et~al., 2013]{WengBusaFeketeHullermeier13}
Weng, P., Busa-Fekete, R., and H{\"{u}}llermeier, E. (2013).
\newblock {I}nteractive {Q}-learning with ordinal rewards and unreliable tutor.
\newblock In {\em {W}orkshop Reinforcement Learning with Generalized Feedback,
  {ECML}/{PKDD}}.

\bibitem[Werbos, 1990]{Werbos90}
Werbos, P.~J. (1990).
\newblock Consistency of hdp applied to a simple reinforcement learning
  problem.
\newblock {\em Neural Networks}, 3:179--189.

\bibitem[Wierstra et~al.,
  2014]{WierstraSchaulGlasmachersSunPetersSchmidhuber14}
Wierstra, D., Schaul, T., Glasmachers, T., Sun, Y., Peters, J., and
  Schmidhuber, J. (2014).
\newblock Natural evolution strategies.
\newblock {\em JMLR}, 15:949--980.

\bibitem[Williams, 1992]{Williams-ml92}
Williams, R. (1992).
\newblock Simple statistical gradient-following algorithms for connectionnist
  reinforcement learning.
\newblock {\em Machine Learning}, 8(3):229--256.

\bibitem[Wilson et~al., 2007]{WilsonFernRayTadepalli07}
Wilson, A., Fern, A., Ray, S., and Tadepalli, P. (2007).
\newblock Multi-task reinforcement learning: A hierarchical bayesian approach.
\newblock In {\em ICML}.

\bibitem[Wilson et~al., 2012]{WilsonFernTadepalli12}
Wilson, A., Fern, A., and Tadepalli, P. (2012).
\newblock A {B}ayesian approach for policy learning from trajectory preference
  queries.
\newblock In {\em Advances in Neural Information Processing Systems}.

\bibitem[Wirth and Neumann, 2015]{WirthNeumann15}
Wirth, C. and Neumann, G. (2015).
\newblock Model-free preference-based reinforcement learning.
\newblock In {\em EWRL}.

\bibitem[Wu and Tian, 2017]{WuTian17}
Wu, Y. and Tian, Y. (2017).
\newblock Training agent for first-person shooter game with actor-critic
  curriculum learning.
\newblock In {\em ICLR}.

\bibitem[Wulfmeier et~al., 2015]{WulfmeierOndruskaPosner15}
Wulfmeier, M., Ondruska, P., and Posner, I. (2015).
\newblock Maximum entropy deep inverse reinforcement learning.
\newblock In {\em NIPS, Deep Reinforcement Learning Workshop}.

\bibitem[Xu et~al., 2007]{xu2007kernel}
Xu, X., Hu, D., and Lu, X. (2007).
\newblock Kernel-based least squares policy iteration for reinforcement
  learning.
\newblock {\em IEEE Transactions on Neural Networks}, 18(4):973--992.

\bibitem[Yu and Zhang, 2013]{YuZhang13}
Yu, T. and Zhang, Z. (2013).
\newblock Optimal {CPS} control for interconnected power systems based on
  {SARSA} on-policy learning algorithm.
\newblock {\em Power System Protection and Control}, pages 211--216.

\bibitem[Yue et~al., 2012]{YueBroderKleinbergJoachims12}
Yue, Y., Broder, J., Kleinberg, R., and Joachims, T. (2012).
\newblock The k-armed dueling bandits problem.
\newblock {\em Journal of Computer and System Sciences}, 78(5):1538--1556.

\bibitem[Zhao et~al., 2010]{ZhaoChenLeungLai10}
Zhao, Q., Chen, S., Leung, S., and Lai, K. (2010).
\newblock Integration of inventory and transportation decisions in a logistics
  system.
\newblock {\em Transportation Research Part E: Logistics and Transportation
  Review}, 46(6):913--925.

\bibitem[Ziebart et~al., 2010]{ZiebartMaasBagnellDey10}
Ziebart, B., Maas, A., Bagnell, J., and Dey, A. (2010).
\newblock Maximum entropy inverse reinforcement learning.
\newblock In {\em AAAI}.

\end{thebibliography}
\bibliographystyle{apalike}
\tableofcontents
\end{document}